\documentclass[acmsmall, nonacm=true]{acmart}
\AtBeginDocument{%
  }

\setcopyright{acmlicensed}
\copyrightyear{2018}
\acmYear{2018}
\acmDOI{XXXXXXX.XXXXXXX}

\usepackage{array}
\usepackage{amsmath}
\usepackage{booktabs}
\usepackage{adjustbox}
\usepackage{algpseudocode}
\usepackage[ruled,vlined]{algorithm2e}
\usepackage[T1]{fontenc}
\usepackage[utf8]{inputenc}
\usepackage{gensymb}
\begin{document}

\title{FLUID: Training-Free Face De-identification via Latent Identity Substitution}

\author{Jinhyeong Park}
\authornote{Both authors contributed equally to this research.}
\email{hini2245@knu.ac.kr}
\author{Muhammad Shaheryar}
\authornotemark[1]
\email{shaheryar@knu.ac.kr}
\author{Seangmin Lee}
\email{smlee0610@knu.ac.kr}
\author{Jong Taek Lee}
\email{jongtaeklee@knu.ac.kr}
\author{Soon Ki Jung}
\authornote{Corresponding author.}
\email{skjung@knu.ac.kr}
\affiliation{%
  \institution{Kyungpook National University}
  \city{Daegu}
  \country{South Korea}
}
%
%
%

\renewcommand{\shortauthors}{Park et al.}

\begin{abstract}
Current face de-identification methods that replace identifiable cues in the face region with other sacrifices utilities contributing to realism, such as age and gender. To retrieve the damaged realism, we present FLUID ($\textbf{F}$ace de-identification in the $\textbf{L}$atent space via $\textbf{U}$tility-preserving $\textbf{I}$dentity $\textbf{D}$isplacement), a single-input face de-identification framework that directly replaces identity features in the latent space of a pretrained diffusion model without affecting the model's weights. We reinterpret face de-identification as an image editing task in the latent \textit{h}-space of a pretrained unconditional diffusion model. Our framework estimates identity-editing directions through optimization guided by loss functions that encourage attribute preservation while suppressing identity signals. We further introduce both linear and geodesic (tangent-based) editing schemes to effectively navigate the latent manifold. Experiments on CelebA-HQ and FFHQ show that FLUID achieves a superior balance between identity suppression and attribute preservation, outperforming existing de-identification approaches in both qualitative and quantitative evaluations.
\end{abstract}

\ccsdesc[500]{Security and privacy~Social network security and privacy}
\ccsdesc[300]{Computing methodologies~Computer vision}
\ccsdesc[100]{Computing methodologies~Machine learning approaches}

\keywords{Face de-identification, face anonymization, utility preservation, diffusion models, image editing, h-space}


\maketitle

\section{Introduction}
The widespread deployment of face recognition (FR) systems has raised urgent concerns about biometric privacy. Facial images are permanent, uniquely identifiable, and easily captured, making them ideal for surveillance but difficult to protect. In modern digital ecosystems, ranging from public CCTV networks to social media platforms, facial data can be collected, stored, and analyzed at massive scale without an individual’s awareness or consent. Even with legal protections such as the General Data Protection Regulation \cite{gdpr} and India’s Digital Personal Data Protection Act \cite{dpdpa} in place, enforcement is challenging and often lags behind rapid advancements in AI-driven surveillance technologies. As a result, de-identification methods are increasingly needed to protect privacy and mitigate risks such as deepfake misuse \cite{westerlund2019emergence} before visual data are shared. This need becomes especially critical as generative AI enables unprecedented manipulation of facial content, amplifying the potential for misuse if source identities are not adequately protected.

Traditional de-identification techniques (e.g., masking, pixelation, and blurring) \cite{babaguchi2009psychological, ilia2015face} often degrade image quality and fail against modern FR models. Such approaches provide only superficial obfuscation and are easily reversed or bypassed by advanced recognition systems trained on large-scale datasets \cite{radiya-dixit2022data}. Prior work has shown that basic obfuscation techniques and data-poisoning cloaks do not offer long-term or robust privacy, as the results are still in danger of identity leakage \cite{oh2016faceless} and adversaries are reversible \cite{nie2022diffusion}. Recent generative models, such as generative adversarial networks (GANs) \cite{goodfellow2014generative} and diffusion models (DMs)~\cite{ho2020denoising} offer promising alternatives with improved realism~\cite{wen2022identitydp, cai2024disguise, shaheryar2025black}. These models introduce structured transformations capable of removing identity information while preserving visual coherence, producing images that are in high quality. However, while these generative methods successfully achieve more realistic identity suppression than traditional ones, they still suffer from a fundamental limitation: most produce faces that are immediately recognizable as de-identified. This occurs because they often fail to preserve well-known non-identity attributes, such as gender and age, which leads to outputs that appear artificial or uncanny despite being visually artifact-free. This visible alteration introduces a subtle but important privacy paradox. When an image clearly signals that de-identification has been applied, it undermines privacy rather than strengthening it. First, conspicuous modifications explicitly reveal that sensitive information has been removed, which can attract adversarial scrutiny and motivate targeted attempts to reverse or exploit the anonymization, which would be unlikely if the image appeared authentic. Second, detectably de-identified faces reduce the credibility of visual data in applications where both privacy protection and perceptual authenticity are essential, such as medical case reporting, journalistic documentation, legal evidence, and public research datasets. In real-world deployments, these unnatural distortions do not merely indicate that anonymization has occurred; they can also erode trust in the data itself. An effective face de-identification method should therefore go beyond identity suppression and preserve key semantic cues, producing outputs that remain perceptually indistinguishable from natural, unmodified faces. Only such realism can prevent adversarial suspicion while maintaining the authenticity required for practical, high-stakes data sharing.

To address these limitations, we turn to image editing in the latent space, hypothesizing its superior attribute preservation compared to generative reconstruction approaches. Unlike full image regeneration which often forces a model to re-synthesize identity-irrelevant attributes from scratch, latent editing directly modifies structured internal representations that already encode fine-grained semantic details of the input. Image editing in the latent space operates with semantically disentangled representations, allowing independent and gradual manipulation from the original data. This enables fine-grained control over the input while reflecting required changes. Importantly, such latent-level operations minimize unnecessary modification, making them well-suited for scenarios where utility attributes (e.g., age, gender, emotion, hairstyle) must remain visually consistent. Recent discoveries in the latent spaces of generative models have enabled such training-free semantic image editing, where inverting input images into the latent space and editing by traversing semantic directions (e.g., emotion, color, pose) yield desired results \cite{harkonen2020ganspace, shen2020interfacegan, kwon2022diffusion}. These findings collectively suggest that latent spaces possess strong linearity and smooth semantic geometry, allowing meaningful edits without requiring model retraining. However, current face image editing methods primarily focus on attribute manipulation while preserving identity, making them unsuitable for de-identification. Moreover, prior editing often relies on inadequate prompts, model fine-tuning, or complex pre-processing pipelines \cite{kim2022diffusionclip, avrahami2022blended, zhu2023boundary, sun2024diffam}. Thus, despite the promise of latent structures, existing pipelines are not directly designed for privacy protection and often lack mechanisms to enforce identity removal while retaining facial utility.

To enable realistic de‑identification with strict attribute preservation, we propose FLUID (Face de‑identification in the Latent space via Utility‑preserving Identity Displacement), which performs identity substitution directly in the latent space of a pretrained unconditional DM without the need to tune model weights. Our core idea is to treat identity removal as a controlled displacement problem in a well-structured semantic manifold, rather than a full generative reconstruction task. This space, named \textit{h}-space, of unconditional DMs has been shown to provide a linear, robust, and homogeneous semantic space \cite{kwon2022diffusion}. In this framework, we introduce a set of loss functions to control the outcome of this process, specifically targeting identity suppression and attribute preservation. These losses collectively enforce a disentangled optimization path: identity information is pushed away, while non-identity features are explicitly constrained to remain stable. As a result, our method aims to generate de-identified faces that remain visually indistinguishable from authentic, unmodified images, achieving what we term `covert privacy protection'. By producing outputs that do not visibly signal the application of anonymization, the method avoids attracting adversarial scrutiny and preserves data credibility in scenarios where both privacy protection and perceptual authenticity are essential.

Additionally, we introduce a geodesic (tangent-based) displacement strategy and explore both linear and geodesic edits in latent space, enabling us to examine the trade-off between de-identification strength and attribute retention. This dual-editing view allows us to empirically characterize how different trajectory geometries in latent space influence the final privacy-utility balance. We further assess the realism of results with balanced identity-utility trade-off quantaitatively and qualitatively. To the best of our knowledge, we are the first to perform face de-identification in the latent space of a DM. Our contributions are as follows:

\begin{itemize}
    \item We introduce a face de-identification framework in the \textit{h}-space of an unconditional DM.
    \item We present and compare our tangent-space editing method with the common linear edit, demonstrating a strong balance between de-identification and attribute fidelity.
    \item We demonstrate FLUID's improved performance over SOTA methods in both realism and utility preservation through extensive experiments.
\end{itemize}

\begin{figure*}
    \centering
    \includegraphics[width=\linewidth]{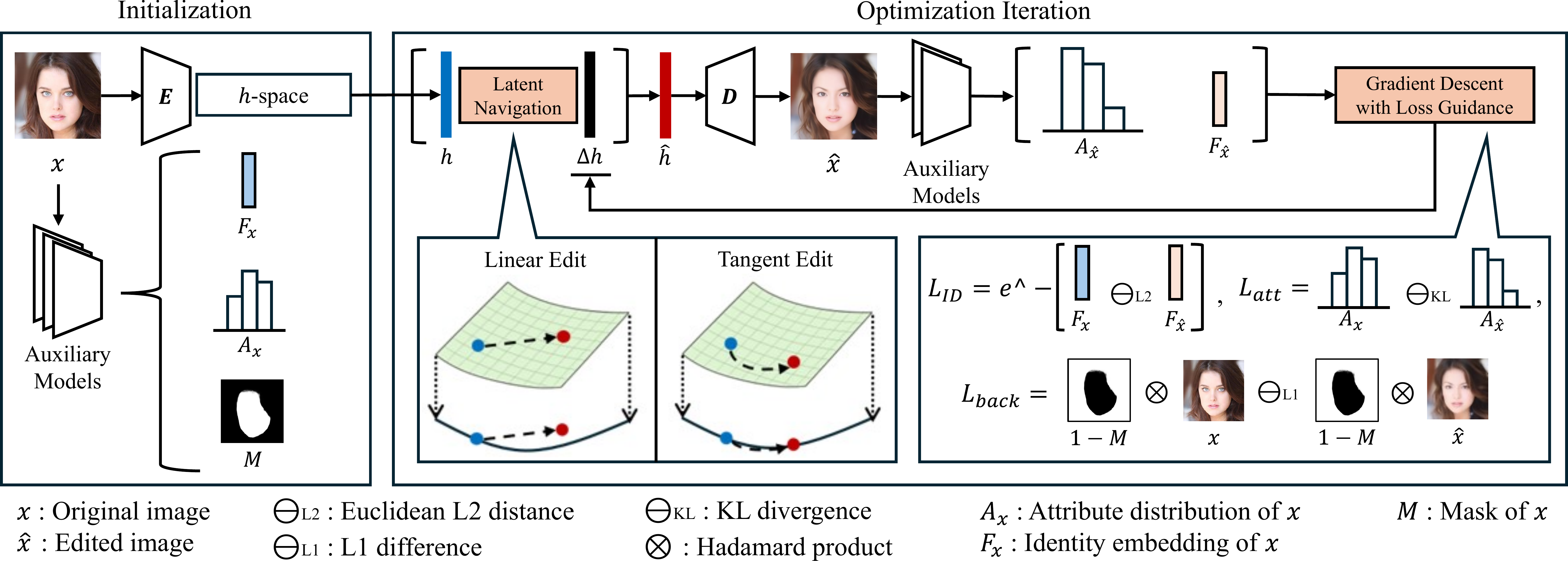}
    \caption{An illustration of our framework. In the initialization procedure, $x$ is inverted into the latent space of a DM to obtain a starting point $h$, a latent vector in the $h$-space, while three auxiliary models are used to get $F_x$, $A_x$, and $M$. During optimization, a direction vector $\Delta h$ is iteratively guided by three loss functions to compare $x$ and $\hat{x}$: an identity loss $L_{ID}$, an attribute preservation loss $L_{att}$, and a face mask loss $L_{mask}$. Decoding $\hat{h}$, the combination of $h$ and $\Delta h$, by either linear or tangent edit results in an edited image $\hat{x}$. Note that $\Delta h$ is initialized randomly with a small norm in the first optimization step.}
    \label{fig:framework}
\end{figure*}

\section{Related Works}
\subsection{Facial De-identification.}
Traditional privacy methods, including blurring and masking \cite{babaguchi2009psychological, ilia2015face}, provide the best efficiency with simple operations but compromise utility and realism. Recent approaches leverage generative models for semantic de-identification, falling into two broad categories: \textit{perturbation-based} and \textit{semantic identity-shifting}. Perturbation-based methods, for example, Fawkes~\cite{shan2020fawkes}, AMT-GAN~\cite{hu2022protecting}, and DiffAM~\cite{sun2024diffam}, introduce adversarial noise or stylized makeup to deceive FR models with subtle changes but maintain perceptual identity, limiting protection against human re-identification. To protect privacy against both humans and FR systems, semantic methods like DeepPrivacy2~\cite{hukkelaas2023deepprivacy2}, Repaint~\cite{lugmayr2022repaint}, and IDDiffuse~\cite{shaheryar2024iddiffuse} replace facial identity altogether, improving both privacy and realism. Though face-swapping approaches such as DCFace \cite{kim2023dcface} and DiffSwap \cite{zhao2023diffswap} can perform semantic shift-based de-identification by giving a synthetic identity as target, they require additional identity inputs, making them vulnerable to misuse in deepfake scenarios where a real identity is given to mimic~\cite{westerlund2019emergence}. Current SOTA semantic shift-based approaches primarily compete and focus on increasing identity dissimilarities between original and de-identified images. While recent advances in generative models have greatly reduced visual artifacts, this aggressive suppression often compromises realism, leading to unintuitive changes in utilities such as age and gender. We argue that the pursuit of extreme identity removal in these approaches sacrifices key utilities, particularly age and gender, thereby undermining the naturalness of de-identified faces that generative methods were originally introduced to improve over traditional techniques. An ideal face de-identification method should instead produce realistic outputs that remain indistinguishable and prevent doubts with further attempts to reverse the de-identification. We tackle this common limitation with a single input without altering the generative model's weights, while most of the methods require training and additional inputs.

\subsection{Face Image Editing.}
Latent space editing in generative models enables facial attribute control without retraining. Early GAN-based methods like InterfaceGAN~\cite{shen2020interfacegan} and StyleFlow~\cite{abdal2021styleflow} identified interpretable directions for attributes (e.g., age, smile), while StyleSpace~\cite{wu2021stylespace} provided fine-grained controls within StyleGAN~\cite{karras2019style}. In DMs, latent traversal has been explored via \textit{h}-space in Asyrp~\cite{kwon2022diffusion}, Boundary Diffusion~\cite{zhu2023boundary}, and semantic latent discovery~\cite{haas2024discovering}. However, these works focus on attribute editing with identity preservation. Our work inverts this paradigm, proposing identity-shifting via optimization in the \textit{h}-space of an unconditional DM preserving non-identity attributes while removing identity. Unlike previous works, we treat identity as an instance-specific component and show it can be decoupled via direction optimization. Our reagent loss-guided editing balances attribute consistency and identity suppression, without retraining.

\section{Preliminaries}
\label{gen_inst}

\subsection{Diffusion Models.}
DMs generate data through a learned denoising process, modeled as a Markov chain over $T$ timesteps. The forward process gradually corrupts clean data $x_0$ with Gaussian noise:
\begin{equation}
    q(x_t|x_{t-1}) = \mathcal{N}(x_t; \sqrt{1 - \beta_t} x_{t-1}, \beta_t \mathbf{I}),
\end{equation}
where $\beta_t \in (0,1)$ is the variance schedule. The reverse process, parameterized by a U-Net \cite{ronneberger2015u} $\epsilon_\theta$, aims to recover clean samples from noise:
\begin{equation}
    p_\theta(x_{t-1}|x_t) = \mathcal{N}(x_{t-1}; \mu_\theta(x_t, t), \Sigma_\theta(x_t, t)).
\end{equation}
The mean is computed as:
\begin{equation}
\mu_\theta(x_t, t) = \frac{1}{\sqrt{\alpha_t}} \left( x_t - \frac{\beta_t}{\sqrt{1 - \bar{\alpha}_t}} \epsilon_\theta(x_t, t) \right),  
\end{equation}
where $\alpha_t = 1 - \beta_t$ and $\bar{\alpha}_t = \prod_{i=1}^t \alpha_i$. The noise predictor $\epsilon_\theta$ is a time-conditioned U-Net trained to estimate the added noise. The iterative denoising process enables precise pixel-level synthesis and surpasses GANs in image quality and diversity \cite{dhariwal2021diffusion, stypulkowski2024diffused}.

\subsection{The \textit{h}-Space.}
Asyrp \cite{kwon2022diffusion} identified that the bottleneck layer of the U-Net in DMs captures a semantic latent space termed \textit{h}-space, which exhibits linearity and disentanglement. This space enables effective image manipulation without retraining the model. Editing is performed by modifying the latent $h_t$ at each timestep with a semantic direction $\Delta h$:
\begin{equation}
    x_{t-1} = \sqrt{\alpha_{t-1}} \mathbf{P}_t(\epsilon_\theta(x_t | \tilde{h}_t)) + \mathbf{D}_t(\epsilon_\theta(x_t)) + \sigma_t z_t,
\end{equation}
where $\tilde{h}_t = h_t + \Delta h$, $\sigma_t$ is the noise scale, and $z_t \sim \mathcal{N}(0, \mathbf{I})$. The transformation $\mathbf{P}_t$ shifts the $x_0$ prediction based on $\tilde{h}_t$, while $\mathbf{D}_t$ reconstructs the path to $x_t$. The key challenge is discovering semantically meaningful directions $\Delta h$ that enable targeted edits without degrading unrelated features. We perform face de-identification in this space than GAN's latent spaces for two reasons: First, GAN inversion is far from perfect due to its significantly reduced dimensionality from pixel space to latent space \cite{pan2023effective}, leading to inaccurate reconstruction. Second, because of the inaccurate GAN inversion, making this inversion accurate requires additional steps, such as fine-tuning the model with a given image \cite{shamshad2023clip2protect} or introducing an additional encoder \cite{maximov2020ciagan}, which increases the complexity of the overall image editing process. As diffusion inversion already offers faithful reconstruction and the \textit{h}-space serves as a semantically meaningful representation, we adopt this space using DDIM inversion \cite{song2020denoising} for effective identity editing.

\section{Proposed Method}
This section introduces our training-free face de-identification using diffusion models (DMs) through latent identity substitution. We remove identity information from the original image in the latent space via loss-driven optimization, then substitute a de-identified representation. Our approach operates entirely within the h-space of a pretrained unconditional DM, requiring no additional training. It consists of three key components: latent inversion, iterative optimization of identity-editing directions via differentiable losses, and reconstruction through the DM's decoder. Figure \ref{fig:framework} provides an overview of our approach, illustrating the initialization procedure where a source image is inverted into the latent space, followed by the optimization process that uses reagent losses to guide identity substitution while preserving other facial attributes. Unlike direct latent replacement, our method discovers identity-specific edit directions through optimization rather than assuming prior knowledge of an identity subspace. 
\subsection{Problem Definition}

Given a facial image \( x \in \mathbb{R}^{H \times W \times 3} \), our objective is to generate a semantically consistent de-identified image \( \hat{x} \) such that the identity difference between \( x \) and \( \hat{x} \) is maximized and the realism perceived by humans is preserved, i.e.,
\begin{equation}
  \hat x
  = \operatorname*{arg\,max}_{\!\tilde x}
    (D_{\mathrm{id}}(x,\tilde x) - D_{\mathrm{human}}(x,\tilde x)),
\end{equation}

where \( D_{\text{id}}(\cdot, \cdot) \) is an identity distance function between two images and \( D_{\text{human}}(\cdot, \cdot) \) is a realism metric determined by human perception.
Instead of manipulating the image directly in pixel space \cite{zhu2023boundary, shen2020interfacegan}, we perform identity editing in the \( h \)-space of a pretrained DM. Let \( f_{\text{inv}} \) be an inversion function that maps the image to its latent representation:
\begin{equation}
    z = f_{\text{inv}}(x), \quad z \in \mathbb{R}^d.
    \label{eq:inversion}
\end{equation}
Our goal is to discover a direction \( \Delta h_{\text{id}} \in \mathbb{R}^d \) in \( h \)-space such that applying this direction to \( z \) suppresses identity-related information:
\begin{equation}
    \hat{z} = z + \lambda \cdot \Delta h_{\text{id}},
    \label{eq:linear}
\end{equation}
where \( \lambda \in \mathbb{R} \) controls the magnitude of editing. Finally  the de-identified image is reconstructed using the DM decoder:
\begin{equation}
    \hat{x} = \mathcal{D}(\hat{z}).
    \label{eq:decode}
\end{equation}
To discover identity-specific directions in the \textit{h}-space, we optimize direction vectors guided by differentiable loss signals. Our experimental results support that face de-identification can be achieved in latent space without degrading other facial attributes. This observation aligns with empirical findings in recent work, such as semantic disentanglement, linear editability, and stable decoder mappings in the bottleneck features of diffusion U-Nets \cite{kwon2022diffusion, jeong2024training}. Building on these properties, we formalize the notion of latent editability and show that identity suppression and substitution can be effectively accomplished by optimizing a loss in latent space without retraining or fine-tuning the diffusion backbone.

\subsection{Latent Navigation: From Euclidean to Geodesic}

To achieve fine-grained identity editing in $\textit{h}$-space, we implement two strategies: linear edits and tangent edits. These strategies define how the latent code $z \in \mathbb{R}^d$ is updated during optimization using a learnable identity-editing direction $\Delta h_{\text{id}}$.

\paragraph{Linear Edit.}
Linear editing is the simplest approach, where identity information is modified by directly adding the editing direction to the original latent, as expressed in Equation \ref{eq:linear}. This formulation is widely used in latent editing literature \cite{kwon2022diffusion, zhu2023boundary}, and is particularly effective for inducing large semantic shifts, such as face de-identification.

\paragraph{Tangent Edit.}
We observe that the manifold of the \textit{h}-space forms a thin hypershell that closely resembles a hypershpere. Although the \textit{h}-space appears to support linear edibility, we hypothesize that this behavior arises because subsequent normalization operations force linearly edited vectors back to the manifold. However, linear edits that momentarily escape the manifold without any consideration of its intrinsic geometry, introduce undesirable artifacts that persist even after normalization. Simply normalizing the linear edited latent to match the original norm (i.e., $||z||$) remains unreliable, as this procedure still inherits the distortions cause by  `cutting through the manifold'. Therefore, to maintain consistency and to appreciate the latent data manifold, we implement a tangent-space traversal strategy, inspired by recent work on geodesic editing in generative latent spaces \cite{hahm2024isometric, jeong2024training}. Rather than applying direct linear addition, we first project identity-editing direction $\Delta h_{\text{id}}$ onto the tangent plane of the hypersphere at $z$ to enable edits remain on the surface of the manifold.
Let $z^{\text{unit}} = z / \|z\|$ and $\Delta h^{\text{unit}}_{\text{id}} = \Delta h_{\text{id}} / \|\Delta h_{\text{id}}\|$. The projection becomes:
\begin{equation}
    \Delta h_{\text{proj}} = \Delta h^{\text{unit}}_{\text{id}} - \langle \Delta h^{\text{unit}}_{\text{id}}, z^{\text{unit}} \rangle \cdot z^{\text{unit}}.
    \label{eq:tangent1}
\end{equation}
Then, the geodesically traversed latent is defined as an automated spherical interpolation (slerp). To automatically perform slerp, we start with the basic form:
\begin{equation}
	\text{slerp}(p_0,p_1;\Omega)=\frac{\text{sin}((1-\Omega)\cdot \Psi)}{\text{sin}(\Psi)} \cdot p_0+\frac{\text{sin}(\Omega \cdot \Psi)}{\text{sin}(\Psi)} \cdot p_1,
\end{equation}
where $p_0$ and $p_1$ are the two points on the arc, $\Psi$ is the angle between $p_0$ and $p_1$, and $\Omega$ is the degree of interpolation between $p_0$ and $p_1$. Since the two vectors $z^{\text{unit}}$ and $\Delta h_{\text{proj}}$ are perpendicular to each other, $\Psi = 90$. Therefore,
\begin{equation}
\begin{aligned}
	\text{slerp}(z^{\text{unit}}, \Delta h_{\text{proj}}; \Omega) 
	&= \frac{\text{sin}((1-\Omega)\cdot 90\degree)}{\text{sin}(90\degree)} \cdot z^{\text{unit}}+\frac{\text{sin}(\Omega \cdot 90\degree)}{\text{sin}(90\degree)} \cdot \Delta h_{\text{proj}}\\
	&=\text{sin}(90\degree - \Omega \cdot 90\degree) \cdot z^{\text{unit}} + \text{sin}(\Omega \cdot 90\degree) \cdot \Delta h_{\text{proj}}\\
	&=\text{cos}(\Omega \cdot 90\degree) \cdot z^{\text{unit}} + \text{sin}(\Omega \cdot 90\degree) \cdot \Delta h_{\text{proj}}.
\end{aligned}
\end{equation}
Finally, with $\theta = \Omega \cdot 90\degree$, we have our final optimization objective as:

\begin{equation}
    \hat{z} = \|z\| \cdot \left( \cos(\theta) \cdot z^{\text{unit}} + \sin(\theta) \cdot \Delta h_{\text{proj}} \right).
    \label{eq:tangent2}
\end{equation}
Here, we substitute $\theta$ with an angle given by the norm of $h_{id}$ in radians. This way, we can still guide the latent walk's direction and intensity, consider the geodesic manifold, and eliminate a hyperparameter (i.e., editing strength $\lambda$) at the same time. Linear edits offer simplicity, while tangent edits preserve manifold geometry for more localized and realistic transformations.

\begin{algorithm}[t]
\small
\caption{Flow of FLUID}
\label{alg:fluid}

\textbf{Input:} Input image $x$, pretrained decoder $D$, weights $\lambda_{\text{id}}, \lambda_{\text{attr}}, \lambda_{\text{mask}}$, step size $\zeta$, editing type (\texttt{linear} or \texttt{tangent}), initialized $\Delta h_{\text{id}}$'s norm $m$, number of optimization steps $n_{\text{opt}}$.

\textbf{Initialize:}

$z \gets f_{\text{inv}}(x)$ \Comment{Equation \ref{eq:inversion}}

$\Delta h_{\text{id}} \sim \mathcal{N}(0, I) \times m$ 

\textbf{Optimization:}

\For{$n \gets 1$ \KwTo $n_{\text{opt}}$}{

    \If{editing type is \texttt{linear}}{
        $\hat{z} \gets z + \lambda \cdot \Delta h_{\text{id}}$ \Comment{Equation \ref{eq:linear}}
    }
    \Else{
        $z_{\text{unit}} \gets z / \|z\|$, \quad $\Delta h_{\text{unit}} \gets \Delta h_{\text{id}} / \|\Delta h_{\text{id}}\|$ 

        $\Delta h_{\text{proj}} \gets \Delta h_{\text{unit}} - \langle \Delta h_{\text{unit}}, z_{\text{unit}} \rangle \cdot z_{\text{unit}}$
        \Comment{Equation \ref{eq:tangent1}}

        $\hat{z} \gets \|z\| \cdot \left( \cos(\theta) \cdot z_{\text{unit}} + \sin(\theta) \cdot \Delta h_{\text{proj}} \right)$
        \Comment{Equation \ref{eq:tangent2}}
    }

    $\hat{x} \gets D(\hat{z})$ \Comment{Equation \ref{eq:decode}}

    $\mathcal{L}_{\text{total}} = \lambda_{\text{id}} \mathcal{L}_{\text{id}} + \lambda_{\text{attr}} \mathcal{L}_{\text{attr}} + \lambda_{\text{mask}} \mathcal{L}_{\text{mask}}$ \Comment{Equation \ref{eq:lfinal}}

    $\Delta h_{\text{id}} \gets \Delta h_{\text{id}} - \zeta \cdot \nabla_{\Delta h_{\text{id}}} \mathcal{L}_{\text{total}}$
}

\textbf{return} De-identified image $\hat{x}$
\end{algorithm}

\subsection{Loss Functions}
To guide this latent substitution process, we introduce a set of differentiable loss functions that steer the optimization toward a desired transformation. These loss functions form a tunable and composable framework that enables selective identity editing in the latent space while preserving the overall utility and semantic consistency of the reconstructed image.

\paragraph{Identity Loss.}
To ensure effective suppression of identity information, we introduce an identity loss based on FaceNet \cite{schroff2015facenet}, a widely used FR model. The model encodes images into an embedding space where Euclidean distance correlates with identity difference. Given a source image $x$ and a generated image $\hat{x}$, we define the identity loss as:
\begin{equation}
    \mathcal{L}_{\text{id}} = e^{-D_{\text{id}}(F_{x}, F_{\hat{x}}))},
    \label{eq:identity_loss}
\end{equation}
where $F_{x}$ denotes the FaceNet embedding of image $x$ and $D_{\text{id}}$ is the Euclidean L2 distance. This exponential form encourages increasing dissimilarity between the source and de-identified image in the embedding space.

\paragraph{Attribute Preservation Loss.}
To retain non-identity facial semantics, we adopt an attribute preservation loss using FaceXFormer \cite{narayan2024facexformer}. FaceXFormer offers multiple facial analysis tasks, including attribute estimation, which measures 40 attributes originated from the CelebA dataset \cite{liu2015deep}. We extract the raw attribute probability vectors from both the original and generated images and minimize their KL divergence:
\begin{equation}
    \mathcal{L}_{\text{att}} = D_{\text{KL}}(A_{x} || A_{\hat{x}}) = \sum_{a=1}^{N_{\text{att}}} A_{x}^{(a)} \log \left( \frac{A_{x}^{(a)}}{A_{\hat{x}}^{(a)}} \right),
    \label{eq:attribute_loss}
\end{equation}
where $A_{x}$ and $A_{\hat{x}}$ are the attribute probability distributions predicted by FaceXFormer, and $N_{\text{att}} = 40$.

%

\paragraph{Face Mask Loss.}
To localize identity transformation and reduce artifacts, we apply a masked L1 loss using a facial region mask $M$ extracted via FaceXFormer’s face parser function. Note that we omit the final thresholding step that discretizes each pixel's probability to 0 or 1, and obtain the raw probabilities of pixels which belong to facial components. This helps restrict significant changes to pixels related to face components (e.g., eyebrows, nose, lips) and smoothens the boundary splitting facial components and non-facial components, reducing boundary-like artifacts seen in inpainting-based de-identification approaches \cite{lugmayr2022repaint, hukkelaas2019deepprivacy, hukkelaas2023deepprivacy2}. The loss function can be written as:
\begin{equation}
    \mathcal{L}_{\text{mask}} = \sum_{i} |x_i - \hat{x}_i| (1 - M_i),
    \label{eq:background_loss}
\end{equation}
where $x_i$ and $\hat{x}_i$ are the pixel values at index $i$ in the original and generated images, respectively, and $M_i$ is the probability face mask.

\paragraph{Total Objective.}
The three losses are combined as:
\begin{equation}
    \mathcal{L}_{\text{total}} = \lambda_{id} \mathcal{L}_{\text{id}} + \lambda_{att} \mathcal{L}_{\text{att}} + \lambda_{mask} \mathcal{L}_{\text{mask}},
    \label{eq:lfinal}
\end{equation}
where $\lambda_{id}=1.0$, $\lambda_{att}=1.0$, and $\lambda_{mask}=0.5$ are weights that control the relative importance of each loss term. This weighted formulation enables fine-grained control over the trade-off between semantic fidelity and de-identification strength, offering a principled and flexible strategy for identity editing in a training-free diffusion framework.

\subsection{Procedure of FLUID} 
\begin{figure*}[t]
    \centering
    \includegraphics[width=\textwidth]{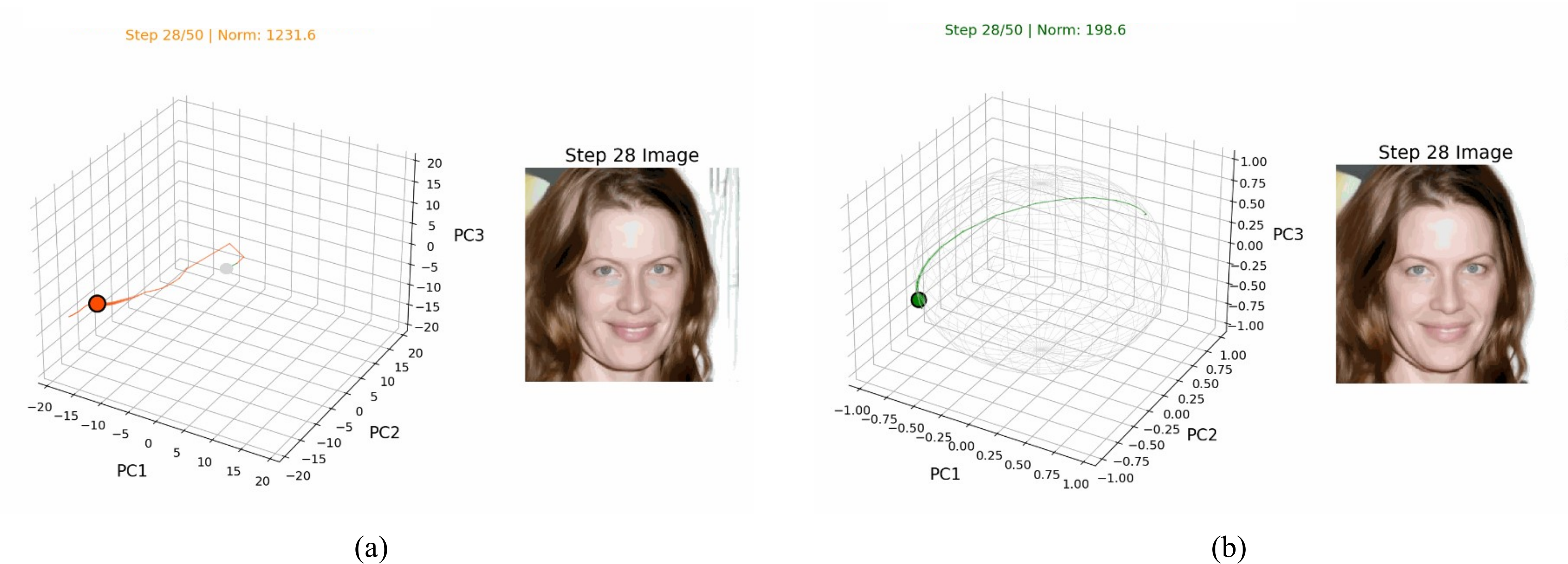}
    \caption{Comparison of linear against tangent edits in the \textit{h}-space. We fit a 3-dimensional principal component analysis (PCA) pool with $5000$ vectors sampled from $100$ editing trajectories, and project the desired vectors onto this PCA space to visualize their trajectory. (a) Linear edits apply raw $\Delta h_{\text{id}}$, allowing the vector norm to drift beyond the typical shell and producing visual artifacts. (b) Tangent edits project $\Delta h_{\text{id}}$ onto the local tangent plane at a constant norm, constraining the trajectory and reducing artifacts.}
    \label{fig:trajectory}
\end{figure*}

Combining the components introduced in the methodology, we summarize FLUID’s overall workflow using pseudocode in Algorithm \ref{alg:fluid}. Our method starts with initialization, where an identity embedding, an attribute distribution, and a face mask are obtained by auxiliary models given an input image. A direction vector $\Delta h_{\text{id}}$ is initialized randomly as a vector with a small norm $m$. Then $\Delta h_{\text{id}}$ and the original image's vector $z$ are fused by either linear edit (Equation \ref{eq:linear}) or a tangent edit (Equations \ref{eq:tangent1} and \ref{eq:tangent2}), producing an edited latent vector $\hat{z}$. The optimization steps proceed by repeatedly calculating losses between the input images and the edited image (Equation \ref{eq:lfinal}), made by decoding $\hat{z}$ (Equation \ref{eq:decode}). These loss values are backpropagated to update $\Delta h_{\text{id}}$, progressively refining the identity-editing direction. After a designated number of steps $n_{opt}$, the final latent vector $\hat{z}$ is decoded through the DM's denoising process, yielding a de-identified image $\hat{x}$. Inspired by spherical interpolation’s success in other generative models, we therefore traverse edits along local tangent planes, controlling the angular step size to stay within the observed shell. As Figure \ref{fig:trajectory} and Table \ref{tab:comparison} show, this angular traversal yields fewer artifacts and less utility damage than straight linear edits, which may drift beyond the typical norm range into regions where the decoder produces unrealistic or unstable outputs.

\section{Experiments}
\label{sec:experiments}
\subsection{Baselines}
To benchmark the effectiveness of our training-free facial identity editing approach, we compare it against SOTA semantic shift-based face de-identification models. We report FALCO~\cite{barattin2023attribute} \footnote{\footnotesize We report FALCO on CelebA-HQ following the authors’ official evaluation protocol (CelebA-HQ split and attribute-based evaluation). Extending FALCO to other datasets requires dataset-specific engineering and, in some cases, additional dataset information or annotations to enable comparable attribute-based metrics. As such extensions may introduce protocol-dependent variations and raise concerns about inconsistent evaluation, we omit multi-dataset FALCO results. This choice is consistent with prior works~\cite{kung2025face,kung2025nullface,he2024diff,yang2024g}, which also report FALCO primarily on CelebA-HQ.}, DeepPrivacy2~\cite{hukkelaas2023deepprivacy2}, Repaint~\cite{lugmayr2022repaint}, the edit variant of Diffprivate~\cite{le2025diffprivate}, and FAMS \cite{kung2025face}, which directly synthesize alternative identities via generative priors. We use the official implementations and default hyperparameter settings provided in their respective publications for comparison. Our method uses a fixed learning rate of $lr = 0.0005$, an editing strength parameter of $\lambda = 1000$, and a diffusion inversion timestep of $T = 600$ with Adam optimizer \cite{kingma2014adam}. 

\subsection{Datasets}
To evaluate our training-free identity editing framework, we adopt two standard benchmarks: CelebA-HQ~\cite{karras2017progressive} and FFHQ~\cite{karras2019style}. CelebA-HQ offers 30,000 high-resolution celebrity faces with 40 annotated binary attributes, enabling fine-grained assessment of attribute preservation. FFHQ comprises 70,000 diverse, high-quality face images, supporting evaluation under natural variations in age, pose, lighting, and ethnicity. Together, these datasets ensure a comprehensive evaluation across both structured attribute supervision and unconstrained real-world diversity.

\subsection{Evaluation Metrics}
To rigorously assess identity suppression, attribute preservation, and detectability, we adopt the following metrics.

\paragraph{Identity Deviation.}
We define identity deviation as source identity distance (SID), which measures the difference between the original and edited images in terms of identity. To quantify SID while measuring the transferability of de-identification across various FR models, we compute this metric using three popular FR models; ArcFace \cite{deng2019arcface}, SphereFace \cite{liu2017sphereface}, and MobileFaceNet \cite{chen2018mobilefacenets}. SID is measured as the inverse cosine similarity between a pair of face embeddings in the FR feature space, each belonging to the original image $x$ and the de-identified result $\hat{x}$:
\begin{equation}
    R_{\text{id}} = 1 - \cos \left( F_{id}(x), F_{id}(\hat{x}) \right),
    \label{eq:identity_eval}
\end{equation}
where $F_{id}(\cdot)$ denotes the face identity encoder.

\begin{figure*}[t]
    \centering
    \includegraphics[width=1.0\linewidth]{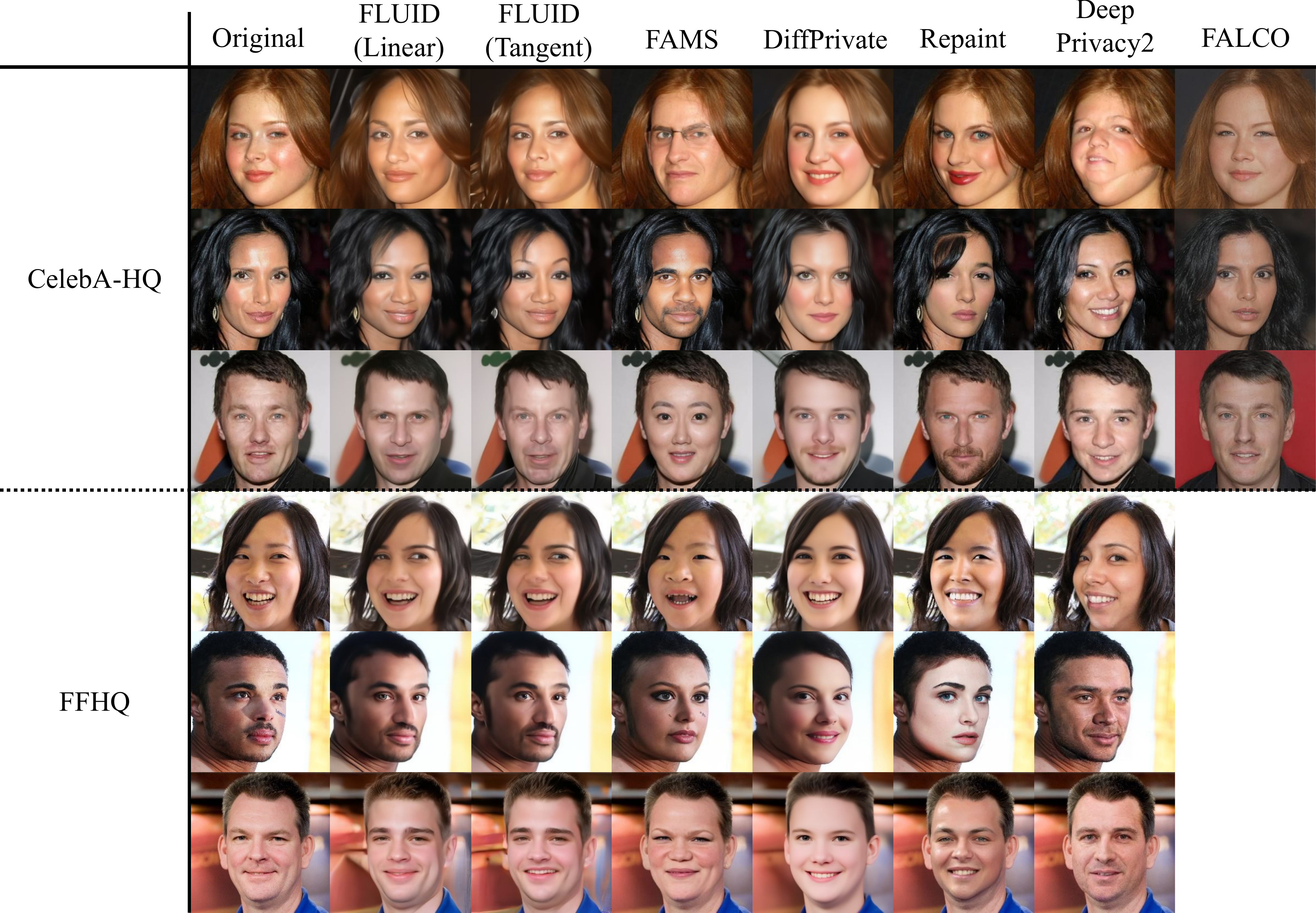}
    \caption{Qualitative comparison with SOTA face de-identification methods on in-domain (CelebA-HQ) and out-of-domain (FFHQ) images. FLUID produces de-identified face images with stricter attribute preservation than other methods. While CelebA-HQ aligns with the pretrained DM’s distribution, FFHQ highlights FLUID’s generalization ability to more diverse real-world inputs.}
    \label{fig:comparison}
\end{figure*}

\paragraph{Attribute Preservation.}

To evaluate semantic consistency for realistic de-identification, we focus on gender and age. Gender preservation is estmated using DeepFace~\cite{serengil2021hyperextended} as:
\begin{equation}
R_{\text{gender}} = \mathbb{I}\left[ \text{DeepFace}(x) = \text{DeepFace}(\hat{x}) \right],
\label{eq:gender}
\end{equation}
where $\mathbb{I}[\cdot]$ is the indicator function. The number of results with gender preserved is counted as gender preservation rate.

Age preservation is measured by FaceXFormer's age estimation function. Specifically, each age range is categorized by age buckets, and we count the age preservation rate just like Equation \ref{eq:gender} as:
\begin{equation}
R_{\text{age}} = \mathbb{I}\left[ \text{FX}_{age}(x) = \text{FX}_{age}(\hat{x}) \right],
\label{eq:age}
\end{equation}
where $\text{FX}_{age}$ stands for FaceXFormer's age estimator.

Often, face image generative models distort the poses to alter the perceived geometry and perform de-identification more easily. We additionally evaluate pose consistency using the pose estimator function from FaceXFormer and compute a deviation-based preservation score that combines pitch, yaw, and roll angular differences into a single metric. Given pose deviations $\Delta_{\text{pitch}}$, $\Delta_{\text{yaw}}$, and $\Delta_{\text{roll}}$ between original and de-identified images, we first calculate the root mean square (RMS):
 \begin{equation}
 \text{RMS}_{\text{pose}} = \sqrt{\frac{\Delta_{\text{pitch}}^2 + \Delta_{\text{yaw}}^2 + \Delta_{\text{roll}}^2}{3}}.
 \label{eq:rms}
 \end{equation}
 The pose preservation score is then normalized as:
 \begin{equation}
 R_{\text{pose}} = \max\left(0, 1 - \frac{\text{RMS}_{\text{pose}}}{\tau}\right),
 \label{eq:pose}
 \end{equation}
 where $\tau =5$ represents the angle threshold in degrees for significant pose deviation. Higher scores indicate better pose preservation, with perfect preservation yielding $R_{\text{pose}} = 1$. To the end, our results show strong preservation of gender and age for realistic face de-identification, and notably, pose is also well preserved despite the absence of pose-specific supervision.

\begin{table*}[t]
  \centering
  \fontsize{6.0}{7.0}\selectfont
  \begin{tabular*}{\textwidth}{@{\extracolsep{\fill}}
      l|
      c c c c c c|c c c c c c
    @{}}
    \toprule
    \multicolumn{1}{c|}{ } & \multicolumn{6}{c|}{CelebA-HQ} & \multicolumn{6}{c}{FFHQ} \\
    \cmidrule{2-7}\cmidrule{8-13}
    Method & SID$\uparrow$ & Detect$\uparrow$ & Gender$\uparrow$ & Age$\uparrow$ & Pose$\uparrow$& Overall$\uparrow$ & SID$\uparrow$& Detect$\uparrow$ & Gender$\uparrow$ & Age$\uparrow$ & Pose$\uparrow$ & Overall$\uparrow$\\
    \midrule
      FALCO		& \textcolor{gray}{0.499} & \textbf{1.000} & \textbf{0.923} & 0.658 & 0.542 & \textcolor{gray}{0.668} & - & - & - & - & - & - \\
      DeepPrivacy2   & 0.730 & \textbf{1.000} & 0.897 & 0.645 & 0.458 & 0.753 & \textcolor{gray}{0.680} & \textbf{1.000} & 0.802 & \textbf{0.610} & \textcolor{gray}{0.366} & \textcolor{gray}{0.691} \\
      Repaint        & 0.764 & 0.999 & 0.871 & 0.623 & 0.605 & 0.794 & 0.748 & \textcolor{gray}{0.998} & 0.721 & 0.470 & 0.440 & 0.703\\
      Diffprivate & 0.845 & \textbf{1.000} & 0.752 & 0.524 & \textcolor{gray}{0.436} & 0.745 & \underline{0.812} & \textbf{1.000} & \textcolor{gray}{0.579} & \textcolor{gray}{0.383} & 0.440 & 0.677\\
      FAMS & \underline{0.856} & \textbf{1.000} & \textcolor{gray}{0.431} & \textcolor{gray}{0.480} & \textbf{0.695} & 0.729 & \underline{0.812} & \textbf{1.000} & 0.642 & 0.496 & \textbf{0.645} & 0.761\\
      FLUID (FaceNet, tangent)   & 0.802 & \textbf{1.000} & 0.907 & \underline{0.686} & \underline{0.674} & \underline{0.835} & 0.756 & \textbf{1.000} & \textbf{0.827} & 0.590 & 0.584 & 0.777\\
      FLUID (FaceNet, linear)  & 0.845 & \textbf{1.000} & 0.883 & \textbf{0.690} & 0.646 & \textbf{0.842} & 0.802 & \textbf{1.000} & 0.799 & \underline{0.596} & 0.562 & \underline{0.786}\\
      FLUID (MagFace, tangent) & 0.825 & \textbf{1.000} & \underline{0.918} & 0.650 & 0.620 & 0.827 & 0.801 & \textbf{1.000} & \underline{0.817} & 0.574 & \underline{0.592} & \textbf{0.789} \\
      FLUID (MagFace, linear) & \textbf{0.863} & \textbf{1.000} & 0.889 & 0.631 & 0.599 & 0.829 & \textbf{0.841} & 0.999 & 0.801 &  0.594 & 0.506 & 0.784\\
    \bottomrule
  \end{tabular*}
    \caption{Quantitative performance comparison between FLUID and SOTA face de-identification methods across five key metrics. These include identity removal (SID), image quality (detect), and preservation of three facial attributes: gender, age, and pose. FLUID achieves improved balance between attribute preservation and de-identification, further enhancing realism beyond other SOTA methods. FLUID's results based on MagFace are described as a replacement analysis under Section \ref{sec:whitebox}.}
    \label{tab:comparison}
\end{table*}

\paragraph{Detectability.}
When evaluating the quality of face images, FID~\cite{heusel2017gans}, FIQ~\cite{terhorst2020ser}, and detectability~\cite{king2009dlib, zhang2016joint} are commonly used metrics. FID measures the similarity between the distributions of generated and real data \cite{maximov2020ciagan, li2023riddle, liu2024adv}, whereas FIQ quantifies the quality of a face image as perceived by face recognition (FR) models \cite{cai2024disguise}. Detectability is often computed as the proportion of images successfully detected by face detection models and is treated as a utility to preserve in anonymization systems \cite{wen2022identitydp, maximov2020ciagan, li2023riddle}. However, as FID is based on an Inception network \cite{szegedy2015going} that is trained on ImageNet \cite{deng2009imagenet} without face images, it has recently been considered inappropriate for assessing face image quality \cite{kim2025visual}. For FIQ, although it is intended to measure face image quality, it largely captures how easily an image can be processed by face recognition (FR) models, rather than how realistic it appears to human observers. As a result, its scores frequently diverge from intuitive notions of visual fidelity. Therefore, to assess visual realism and face detectability, we report detectability scores as our metric against FR models and complement the results with a user study for human-perceived quality. The number of face detection successes $R_{\text{det}}$ is measured by both Dlib \cite{king2009dlib} and MTCNN \cite{zhang2016joint}, and we provide the average detection counts.

\paragraph{Overall Metric.}
Evaluating face de-identification systems requires more than checking whether identity is removed; the method must also preserve facial utility and semantic consistency. Existing metrics, however, often reward models that excel in one dimension while quietly failing in others. To address this imbalance, we introduce a \emph{nested harmonic mean} that captures the overall identity–utility trade-off in a principled way. This is to prevent an aggregate from over‑valuing a configuration that inflates one attribute while degrading another. First, we form a harmonic mean score of three attributes:
\begin{equation}
    \mathrm{HM}_{\text{att}}=\frac{3}{\frac{1}{R_{\text{gender}}}+\frac{1}{R_{\text{age}}}+\frac{1}{R_{\text{pose}}}},
\end{equation}
then apply another harmonic mean with $R_{\text{id}}$, $R_{\text{det}}$, and $\mathrm{HM}_{\text{att}}$ to obtain the overall score $\mathrm{HM}_{\text{overall}}$. Because harmonic mean is dominated by low components, a method cannot mask weak privacy or a degraded attribute with a single strong metric; nesting enforces internal balance among attributes before combining the three pillars (privacy, detectability, semantics) concisely.

\begin{figure}[t]
    \centering
    \includegraphics[width=0.5\columnwidth]{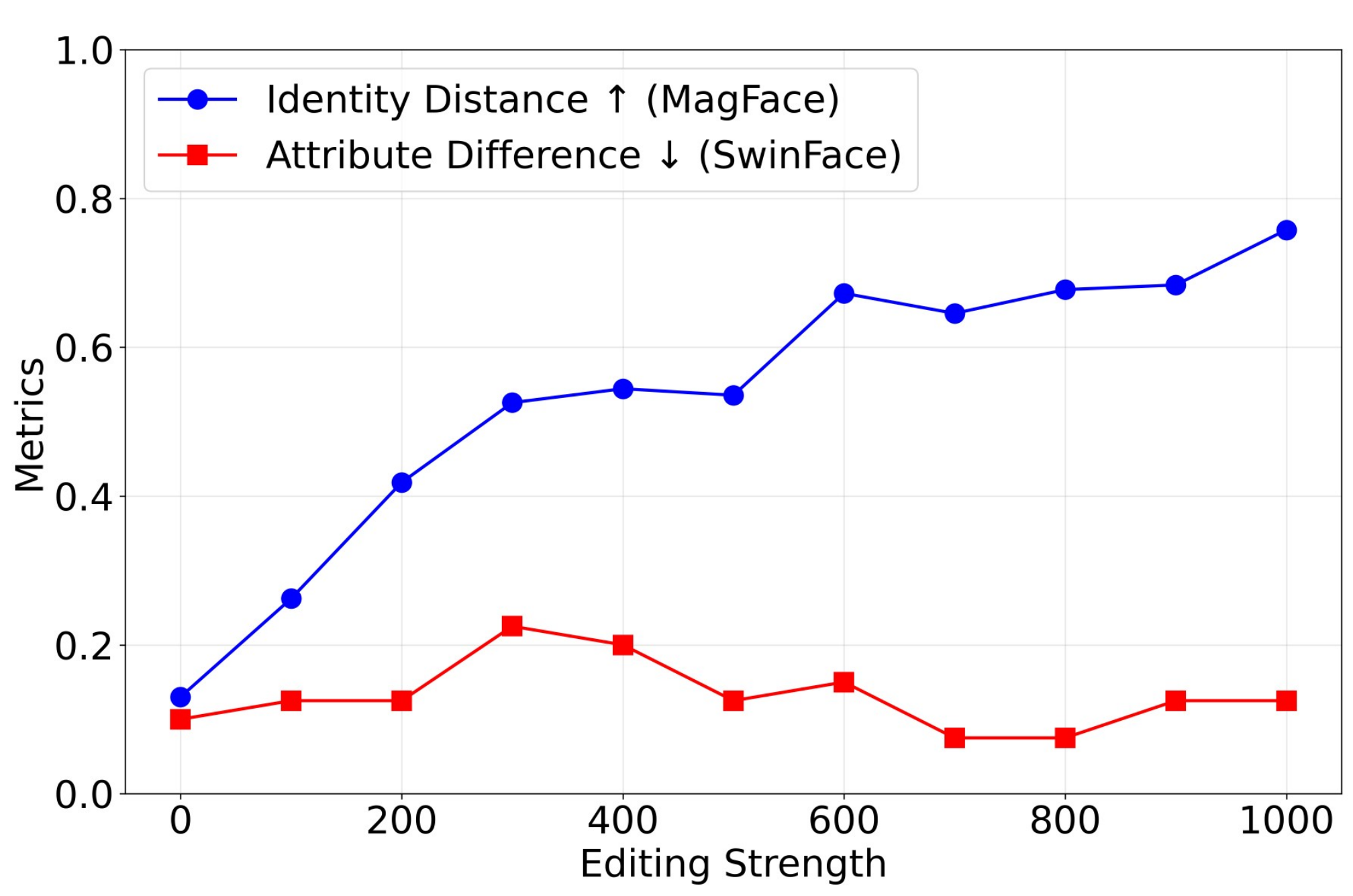}
    \caption{Identity-attribute trade-off curve across increasing linear edit strengths.}
    \label{fig:metric}
\end{figure}

\section{Results}
\paragraph{Quantitative and qualitative comparison.} The closest prior work to ours is FALCO, as both methods optimize a latent vector in the latent space of a generative model with CelebA-defined attributes. However, various differences exist: First, FALCO performs face image de-identification in the popularly used latent space of StyleGAN2 \cite{karras2020analyzing}, while we solely rely on the \textit{h}-space of an unconditional DM for the first time; Second, FALCO is primarily designed for dataset-level anonymization, with the explicit goal of preserving CelebA attribute distributions to maintain dataset utility, rather than for per-image visual fidelity. This design choice is reflected in Table \ref{tab:comparison} and Figure \ref{fig:comparison}, where FALCO achieves the lowest SID, exhibiting noticeable background distortion, and damaging pose preservation by generating a new image rather than editing the original one; Finally, FLUID improves both visual fidelity and the identity-utility trade-off, thereby extending face de-identification beyond dataset-level anonymization to a more general and flexible setting. This is evidenced by similar attribute preservation performances and significantly improved SID in Table \ref{tab:comparison}, and notably better background preservation in Figure \ref{fig:comparison}.\\
On CelebA-HQ, FLUID's tangent edit yields higher gender preservation (0.907) than other models, alongside strong age (0.686) and pose (0.674) consistency. While linear edits show higher SID (0.845) than that of tangent edits (0.802), they often introduce off-manifold artifacts (first and last row of FLUID Linear in Figure \ref{fig:comparison}) and attribute drifts (gender and age pillars in Table \ref{tab:comparison}), which reduce visual realism and semantic stability. Meanwhile, Diffprivate and FAMS achieve higher SID than most of the compared methods by simply sacrificing attributes, producing artifactless but awkward images. Specifically, we observe that Diffprivate tends to smooth the images, making the results femine and younger (FFHQ results of Diffprivate in Figure \ref{fig:comparison}), and FAMS mutually altering both genders and ages (all results of FAMS in Figure \ref{fig:comparison}). Additionally, Diffprivate often attempted to frontalize the faces. All together, these tendencies are evident in the lowest scores of these two methods for gender, age, and pose across both CelebA-HQ and FFHQ. Similarly, as inpainting-based methods, DeepPrivacy2 and Repaint often distort poses (first rows of CelebA-HQ and FFHQ of DeepPrivacy2 results in Figure \ref{fig:comparison}) and offer noticeable borderline artifacts (second rows of CelebA-HQ and FFHQ of Repaint in Figure \ref{fig:comparison}) around the face region, but provide much higher gender and age preservation scores than Diffprivate and FAMS by relying on the knowledge of the trained datasets.

\paragraph{Identity disentanglement validation.} 
We assess whether identity-specific directions can be discovered without explicit supervision. Our optimization finds latent directions $\Delta h_{\text{id}}$ that successfully increase identity distance while maintaining stable non-identity attributes. Figure \ref{fig:metric}  illustrates this validation: as editing strength $\lambda$ increases, identity distance (measured by MagFace \cite{meng2021magface}) grows monotonically, while attribute differences (measured by SwinFace \cite{qin2023swinface}) remain consistent. This empirical evidence supports our core assumption that identity and semantics can be decoupled through loss-guided optimization. 

\paragraph{Tangent vs. linear editing.}
Why does tangent editing outperform linear editing, and does it generalize? In high-dimensional latent spaces, linear edits often drift off the data manifold, resulting in degraded outputs. Tangent editing, by contrast, constrains the edit direction to the local tangent plane of $h$, preserving both norm and curvature (as shown in Equations \ref{eq:tangent1} and \ref{eq:tangent2}). This geodesic formulation avoids semantic drift, significantly improving attribute preservation, visual quality, and cross-domain generalization, particularly in challenging settings such as emotion and pose changes (Table~\ref{tab:comparison}).

\begin{figure}[t]
    \centering
    \includegraphics[width=0.4\columnwidth]{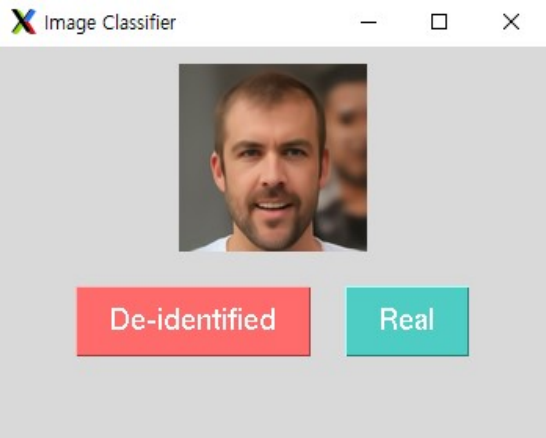}
    \caption{The GUI interface for our user study. Users are asked to decide whether the given images are processed by a de-identification method or not. The image name and source are hidden for fair blind test.}
    \label{fig:gui}
\end{figure}

\begin{figure}[t]
    \centering
    \includegraphics[width=0.7\columnwidth]{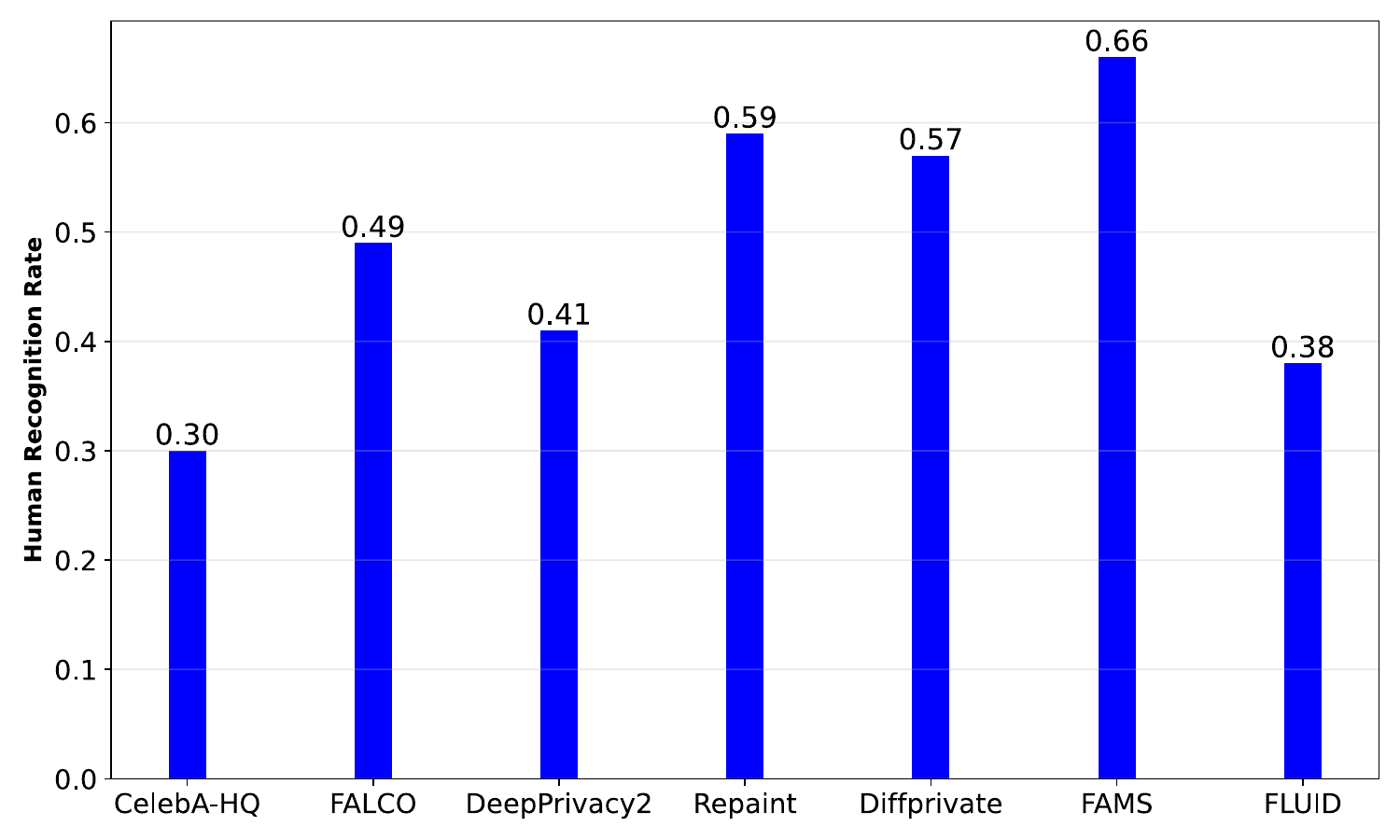}
    \caption{Human recognition of de-identification results based on our user study. FLUID and DeepPrivacy2 are considered the most realistic, showing low recognition rates similar to that of CelebA-HQ.}
    \label{fig:user}
\end{figure}

\begin{figure}[t]
    \centering
    \includegraphics[width=0.6\columnwidth]{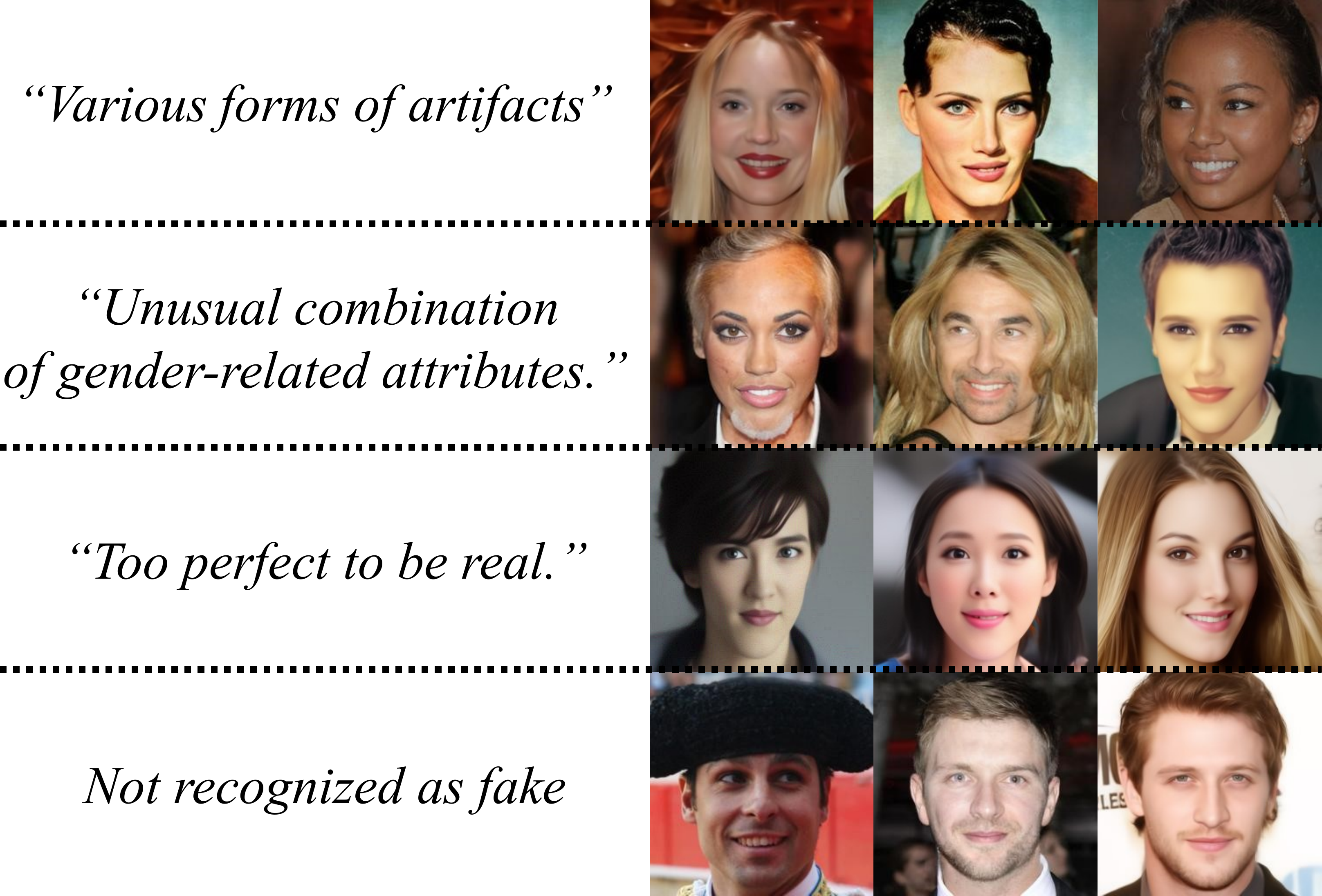}
    \caption{Example images corresponding to each frequently observed user response in our study. Participants frequently recognized low-quality outputs from perturbation-based methods and identified borderline artifacts introduced by inpainting techniques. Additionally, unusual mixtures of gender representations raised user suspicion regarding the authenticity of the data. Notably, the overly perfect appearance of some faces, such as unnaturally clear skin textures and overly smooth hairlines, was also cited as an indicator that the images had been generated or modified by AI models.}
    \label{fig:user_example}
\end{figure}

\begin{figure}[t]
    \centering
    \includegraphics[width=0.6\columnwidth]{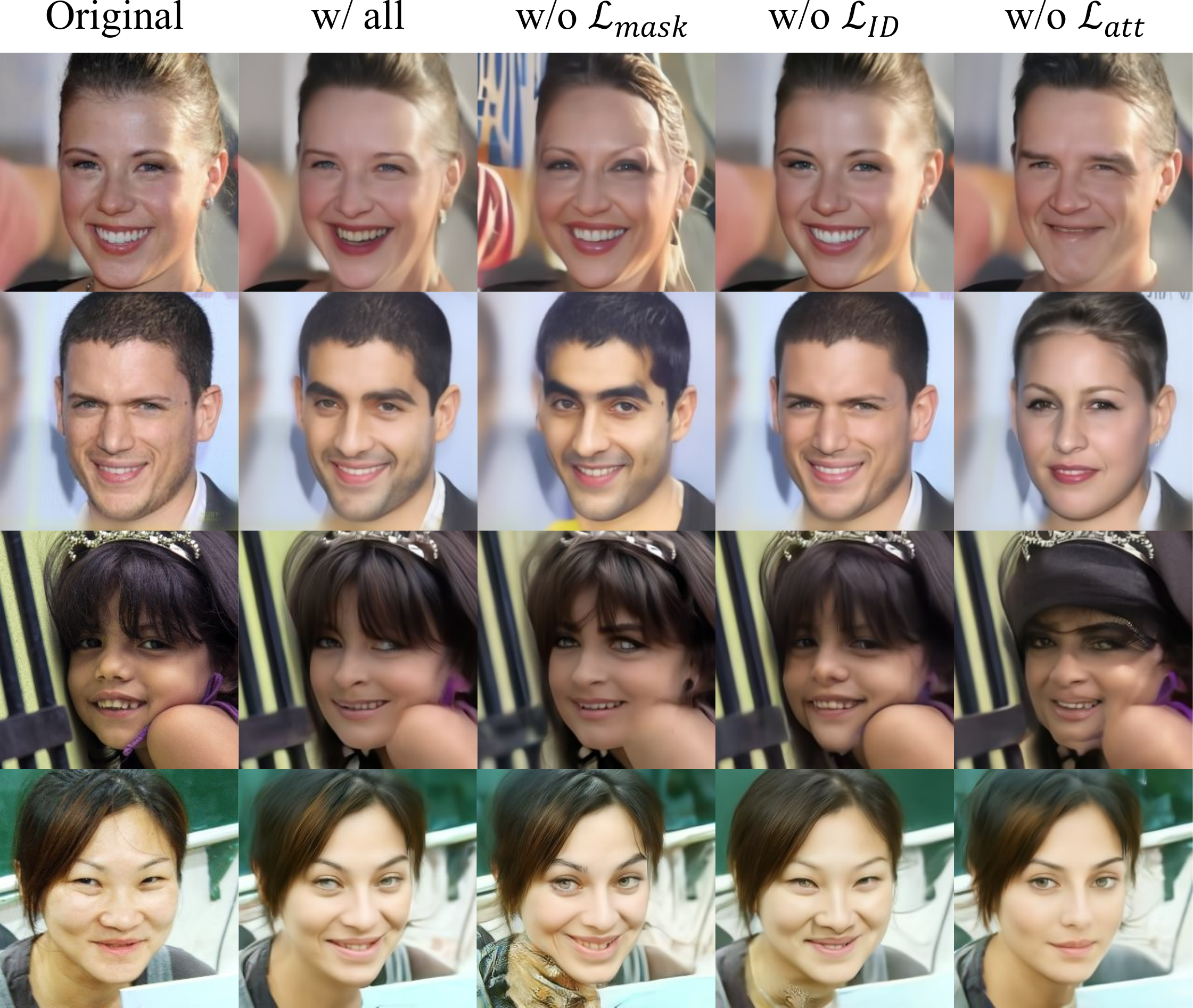}
    \caption{Qualitative results of ablation study for loss function combinations. The top two rows display CelebA-HQ samples, while the bottom rows are based on FFHQ.}
    \label{fig:ablation}
\end{figure}

\subsection{Evaluating Human Recognition on De-identified Faces}
\label{subsec:user_study}
We conducted a user study to evaluate how human perceive de-identified faces. We design a simple graphical user interface (GUI) in Python that shows an image once at a time and asks users to guess whether the image has been de-identified by a generative model or not. We first selected $15$ de-identified images from each compared methods and FLUID, as well as $20$ real images from CelebA-HQ, with all images chosen from those with high SER-FIQ \cite{terhorst2020ser} scores. This is to prevent the users from identifying fake images from mere artifacts generated by models, which could easily guide the users to identify the fakeness without looking at the faces, and automate the user study process to an extent. We recruited $25$ graduate students with diverse nationalities who are not familiar with the field of face de-identification. Note that we do not disclose the number of images selected from each category to the users, nor do we reveal any additional information about the face image that appears in the GUI. An example screenshot of our GUI is shown in Figure \ref{fig:gui}, while the results and example images are illustrated in Figure \ref{fig:user} and \ref{fig:user_example}, respectively.

Besides FLUID achieving the lowest human recognition rate (0.38) among the compared models, counterintuitively, DeepPrivacy2 has attained low human recognition rate as well (0.41) despite its low overall score in Table \ref{tab:comparison}. We attribute this to the following responses collected after the blind test, in which the participants were asked to explain the primary factors that led them to conclude an image was de-identified.

\begin{table}[t]
  \centering
  \fontsize{6.7}{8.5}\selectfont
  \setlength{\tabcolsep}{4.0pt}
  \begin{tabular}{l c c c c c c c}
    \toprule
    Method & SID$\uparrow$ & Detect$\uparrow$
      & Gender$\uparrow$ & Age$\uparrow$ & Pose$\uparrow$ & Overall$\uparrow$\\
    \midrule
      8 steps & 0.897 & 0.999 & 0.873 & 0.677 & 0.610 & 0.848\\
      12 steps & 0.841 & 1.000 & 0.843 & 0.647 & 0.630 & 0.827\\
      16 steps & 0.820 & 1.000 & 0.883 & 0.690 & 0.646 & 0.834\\
      20 steps & 0.788 & 0.999 & 0.876 & 0.669 & 0.681 & 0.824\\
      24 steps & 0.749 & 1.000 & 0.870 & 0.649 & 0.679 & 0.806\\
    \bottomrule
  \end{tabular}
    \caption{Quantitative results on different numbers of denoising steps.}
    \label{tab:steps}
\end{table}


\begin{table}[t]
  \centering
  \fontsize{6.7}{8.5}\selectfont
  \setlength{\tabcolsep}{4.5pt}
  \begin{tabular}{l c c c c c c c}
    \toprule
    $\lambda$ & SID$\uparrow$ & Detect$\uparrow$ 
      & Gender$\uparrow$ & Age$\uparrow$ & Pose$\uparrow$ & Overall$\uparrow$\\
    \midrule
      100  & 0.589 & 1.000 & 0.915 & 0.695 & 0.696 & 0.746\\
      200  & 0.658 & 1.000 & 0.902 & 0.685 & 0.677 & 0.776\\
      300  & 0.669 & 1.000 & 0.904 & 0.664 & 0.671 & 0.777\\
      400  & 0.683 & 0.999 & 0.893 & 0.681 & 0.668 & 0.784\\
      500  & 0.699 & 1.000 & 0.899 & 0.666 & 0.665 & 0.789\\
      600  & 0.736 & 1.000 & 0.880 & 0.669 & 0.646 & 0.800\\
      700  & 0.765 & 1.000 & 0.872 & 0.665& 0.627 & 0.806\\
      800  & 0.790 & 1.000 & 0.870 & 0.671& 0.624 & 0.815\\
      900  & 0.806 & 1.000 & 0.877 & 0.670& 0.640 & 0.824\\
      1000 & 0.820 & 1.000 & 0.883 & 0.690 & 0.646 & 0.834\\
    \bottomrule
  \end{tabular}
  \caption{The impact of linear edit strength $\lambda$ used for linear edits.}
  \label{tab:strength}
\end{table}

\begin{table}[t]
  \centering
  \fontsize{6.7}{8.5}\selectfont
  \setlength{\tabcolsep}{4.5pt}
  \begin{tabular}{l c c c c c c c}
    \toprule
    $n_{step}$ & SID$\uparrow$ & Detect$\uparrow$
      & Gender$\uparrow$ & Pose$\uparrow$ & Age$\uparrow$ & Overall$\uparrow$\\
    \midrule
      10  & 0.568 & 0.999 & 0.924 & 0.711 & 0.742 & 0.743\\
      20  & 0.679 & 0.999 & 0.921 & 0.695 & 0.692 & 0.790\\
      30  & 0.747 & 1.000 & 0.900 & 0.690 & 0.676 & 0.814\\
      40  & 0.800 & 1.000 & 0.890 & 0.685 & 0.657 & 0.829\\
      50  & 0.820 & 1.000 & 0.883 & 0.690 & 0.646 & 0.834\\
      60  & 0.836 & 0.999 & 0.892 & 0.686 & 0.624 & 0.835\\
    \bottomrule
  \end{tabular}
  \caption{The tendency of results based on various numbers of optimization steps.}
  \label{tab:opt_steps}
\end{table}

\begin{table}[h!]
  \centering
  \fontsize{7.5}{8.5}\selectfont
  \begin{tabular}{c|c c c|c c c}
    \toprule
    \multicolumn{1}{c|}{ } & \multicolumn{3}{c|}{CelebA-HQ} & \multicolumn{3}{c}{FFHQ} \\
    \cmidrule{2-4}\cmidrule{5-7}
    Method & Pitch$\downarrow$ & Yaw$\downarrow$& Roll$\downarrow$
      & Pitch$\downarrow$ & Yaw$\downarrow$& Roll$\downarrow$\\
    \midrule
      FALCO        & 2.50 & 2.85 & 1.15 & - & - & - \\
      DeepPrivacy2   & 3.26 & 2.71 & 2.01 & 3.94 & 3.09 & 2.26\\
      Repaint        & 2.89 & 1.55 & 0.97 & \textit{4.25} & 2.00 & 1.22\\
      Diffprivate & 3.24 & 3.52 & 0.99 & 3.49 & 3.24 & 0.92\\
      FAMS           & 2.13 & 1.25 & 0.93 & 2.25 & 1.76 & 1.12 \\
      Ours (tangent)   & 2.39 & 1.58 & 0.80 & 3.07 & 1.91 & 0.99\\
      Ours (linear)  & 2.63 & 1.57 & 0.84 & 3.22 & 1.94 & 1.01\\
    \bottomrule
  \end{tabular}
    \caption{Raw pitch, yaw, and roll angle deviations of each method measured by the pose estimator from FaceXFormer. Pitch angle difference from Repaint under FFHQ shows the greatest deviation of $4.25$.}
    \label{tab:pose}
\end{table}

\paragraph{Quality degradation by various artifacts.} Artifacts from various methods, though trivial in size and shape, encouraged users to determine such images as fake without looking the face region. Notably, users were also able to detect subtle boundaries marked by abrupt color changes across the line, which are commonly observed in mask-based inpainting models such as Repaint and DeepPrivacy2. Beyond such explicit artifacts, users also flagged images with slight blurriness or monotonous background coloration as fake, which the artifacts are frequently produced by FALCO. De-identification methods operating in latent spaces may incompletely transform or reassign original pixel information (e.g., hat to hair, or hair to background), resulting in residual inconsistencies. Many of the detected results from both FLUID and FALCO fall into this category. Consequently, although we intentionally and automatically selected the highest-quality samples across all methods to minimize such cases, typical generation artifacts remained present in the final selection. This observation further indicates that FIQ is not an ideal metric for assessing face image quality from a perceptual perspective, as it fails to filter out artifact-containing images, particularly those outside the facial region.
\paragraph{Mismatching gender information.} Often the de-identification was recognized by unusual mixtures of gender-related attributes. For example, the fusion of common attributes seen from women's images and those from men's (e.g., a bald woman) tend to be identified as fake. Similarly, some users would classify fake images by observing facial skeletal morphologies. In this case, FLUID and DeepPrivacy2 were able to avoid such decisions by showing high gender preservation rates, minimizing the odds of distorting gender information.
\paragraph{Too-perfect images.} DiffPrivate's results were likely to be recognized as de-identified due to perfect-clean skin and hairlines, which are unlikely to be seen in real-world scenarios. FLUID and DeepPrivacy2's results were less recognized as synthesized due to natural skin tones and hair appearances.

\subsection{Ablation Study on Loss Functions}
To evaluate the individual contribution of the proposed reagent loss components, we conduct an ablation study, where we show qualitative examples in Figure~\ref{fig:ablation}. Removing $\mathcal{L}_{\text{ID}}$ removes the objective to displace the original identity, thereby obviously leading to mere reconstruction with trivial changes. Excluding $\mathcal{L}_{\text{att}}$ leads to significant semantic drift, particularly altering key attributes such as gender (top two rows) and facial expression (last row), violating attribute preservation. Other attributes were also discovered to be shifting, such as in the third row where the result without $\mathcal{L}_{\text{att}}$ attempted to add a hat on top of the face. When $\mathcal{L}_{\text{mask}}$ is omitted, we observe varying levels of background artifacts and reduced visual realism. For example, the first and last rows introduce severe artifacts on the background and neck, respectively, while the second and third rows display examples of small portion artifacts on the bottom left region of the images.

\subsection{Influence of Hyperparameter Selection and Analysis}

\paragraph{Denoising steps and computational analysis.}
Similarly, we conduct another comparison to select the number of denoising steps. As written in Table \ref{tab:steps}, increasing the number of reverse steps tends to reduce SID and retain attributes, especially pose. Although 8 denoising steps show the best balance due to the strongest SID, we select 16 steps for our experiments for improved stability and consistency, ensuring reliable performance across diverse inputs and more attribute preservation. Even though the number of denoising steps is much less than that of DDPM thanks to the implementation of DDIM \cite{song2020denoising} inversion and generation, naively backpropagating through the iterative reverse steps still requires considerable amount of memory. Therefore, we implement gradient checkpointing \cite{chen2016training} to significantly reduce memory consumption, enabling us to run the implementation on our hardware, a single Titan XP GPU with 12GB of memory. With the help of gradient checkpointing, FLUID runs with approximately 6GB of memory and takes under 3 seconds for each optimization step, including all auxiliary model operations.

For the duration of injecting the edited \textit{h} vector $\hat{z}$, we mostly follow Asyrp as:
\begin{equation}
p_{\theta}^{(t)}(x_{t-1} \mid x_t) = 
\mathcal{N}\left(\sqrt{\alpha_{t-1}}\, \mathbf{P}_t\left(\epsilon_{\theta}^{t}(x_t)\right) + \mathbf{D}_t, \sigma_t^2 \mathbf{I} \right).
\end{equation}

\noindent
The form of $\epsilon_{\theta}^{t}(\cdot)$, the usage of guidance $\Delta h_{id}$, and the randomness factor $\eta$ depend on timestep $t$:
\begin{itemize}
  \item If $T \geq t \geq t_{\text{edit}}$, the model uses $\eta = 0$ and guided prediction: $\epsilon_{\theta}^{t}(x_t \mid \Delta h_{id})$.
  \item If $t_{\text{edit}} > t \geq t_{\text{boost}}$, the model uses $\eta = 0$ and unconditional prediction: $\epsilon_{\theta}^{t}(x_t)$.
  \item If $t_{\text{boost}} > t$, the model also uses unconditional prediction: $\epsilon_{\theta}^{t}(x_t)$ but $\eta = 1$.
\end{itemize}

Our choice of starting timestep $T=600$ and editing range $t_{\text{edit}}=400$ is motivated by converging evidence from recent work on DM semantics and the critical `mixing step' concept. P2 weighting \cite{choi2022perception} demonstrates that DMs learn perceptually rich semantic content at medium signal-to-noise ratio (SNR) timesteps, while Boundary Diffusion \cite{zhu2023boundary} provides theoretical and empirical evidence for a `mixing step' around $t=500$ where semantic boundaries emerge in the latent space. Through Markov chain mixing time analysis, the authors use $t=500$ in their implementation as the critical convergence point where unconditional DMs begin to exhibit meaningful semantic structure. By restricting our $\Delta h_{\text{id}}$ injection to timesteps [600, 400], we strategically target this semantically rich region that spans the critical mixing step at $t \approx 500$, ensuring that identity-specific transformations occur precisely where semantic boundaries are most well-defined. This approach leverages both the content-learning stage identified by P2 weighting and the semantic boundary formation demonstrated by Boundary Diffusion, enabling more effective identity suppression while maintaining semantic consistency of non-identity facial attributes and avoiding interference with coarse structural learning at higher timesteps. 

On the other hand, our choice of $t_\text{boost}=200$ follows the implementation of Asyrp, which timesteps lower than that enable $\eta = 1$ to avoid additional noise over the resulting image and improve details. Therefore, our denoising strategy consists of edited information injection $(600 \geq t > 400)$, unconditional denoising $(400 \geq t > 200)$, and quality boosting $(200 \geq t > 0)$.

\paragraph{Linear edit strength.} We clarify that the editing strength $\lambda$ for linear edits is set to $1000$. In contrast, Boundary Diffusion offers the editing strength within the range of $[-150, 150]$, as its direction vector encodes the signed distance to a hyperplane (i.e., boundary) defined by a linear SVM. Since we do not rely on such boundaries during optimization, and the norm of $\Delta h_{\text{id}}$ varies throughout the process, we conduct a comparison study on the linear editing strength $\lambda$ in the range of $100$ to $1000$, as shown in Table~\ref{tab:strength}. Unlike other replacement experiments where trade-offs are evident, we observe trivial trade-off patterns in this case. Larger values of $\lambda$ consistently lead to increased SID, with constantly high detectability scores and slightly decreasing attribute preservation, thereby improving the overall score. We attribute the increase in SID to a larger exploration range enabled by stronger scaling factors within a limited number of optimization steps. Based on this observation, we select $\lambda = 1000$ as our default linear edit strength.

\begin{figure*}[t]
    \centering
    \includegraphics[width=\textwidth]{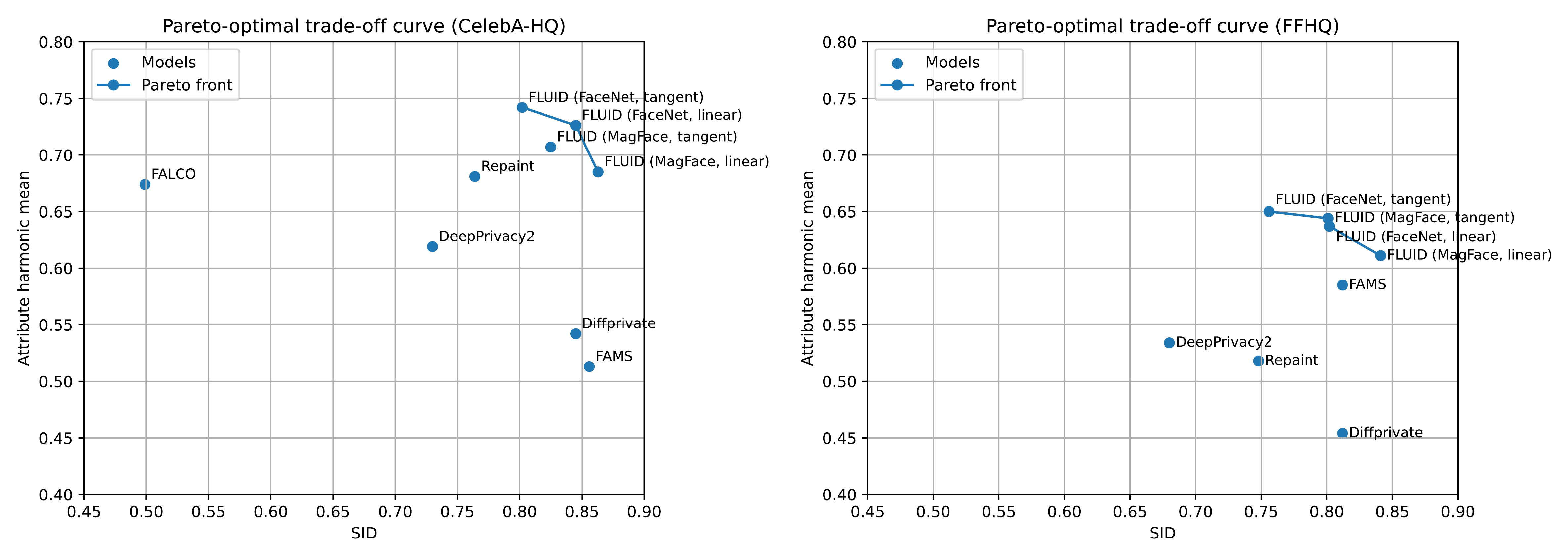}
    \caption{Pareto front curves and data points plotted for each dataset. The Pareto front is formed constantly with FLUID's data points, while other models are located behind the front.}
    \label{fig:pareto}
\end{figure*}

\paragraph{Optimization steps.}
In Table \ref{tab:opt_steps}, we show the impact of the number of optimization steps on the overall performance. The identity-attribute trade-off is also shown here by 10 and 20 steps with low SID and high attribute preservation, indicating insufficiency in de-identification strength and, ultimately, optimization steps. Although 30 steps already outperforms other SOTA methods, we select 50 steps for our experiments due to the high SID at the cost of a minor decrease in attribute preservation rates. While a higher number of steps yields better overall balance and more iterations might further marginally improve, our results show that 50 steps already achieve competitive performance compared to SOTA methods, while results around 40 steps start to plateau.

\subsection{The FR Model's Impact for White-box Attack}
\label{sec:whitebox}
We provide an example when the FR model that is used for white-box attacks is altered. For this purpose, selecting a fundamentally different FR model from FaceNet will yield meaningful outcomes. We replace FaceNet, an L2 distance-based model, with MagFace \cite{meng2021magface}, an angular distance-based FR model where cosine similarity correlates with identity similarity. Since FaceNet and MagFace differs in design (i.e., the distance metric), we introduce a new identity loss function suitable for MagFace. Given a source image $x$ and a generated image $\hat{x}$, we define this identity loss as:
\begin{equation}
    \mathcal{L}_{\text{id}} = |CS(F_{x}, F_{\hat{x}})|,
    \label{eq:identity_loss_relu}
\end{equation}
where $F_{x}$ denotes the MagFace embedding of image $x$ and CS stands for cosine similarity. In Table \ref{tab:comparison}, we append MagFace's performance after that of FaceNet. The improvement over utilities is also shown by MagFace-based linear and tangent edits, aligning with our hypothesis. It is important to note that white-box attacks against MagFace produce the highest SID values and yet higher balance scores compared to other SOTA baselines. However, the age and pose scores are sacrificed compared to FaceNet results, posing the identity-utility trade-off.

\begin{figure*}
    \centering
    \includegraphics[width=\linewidth]{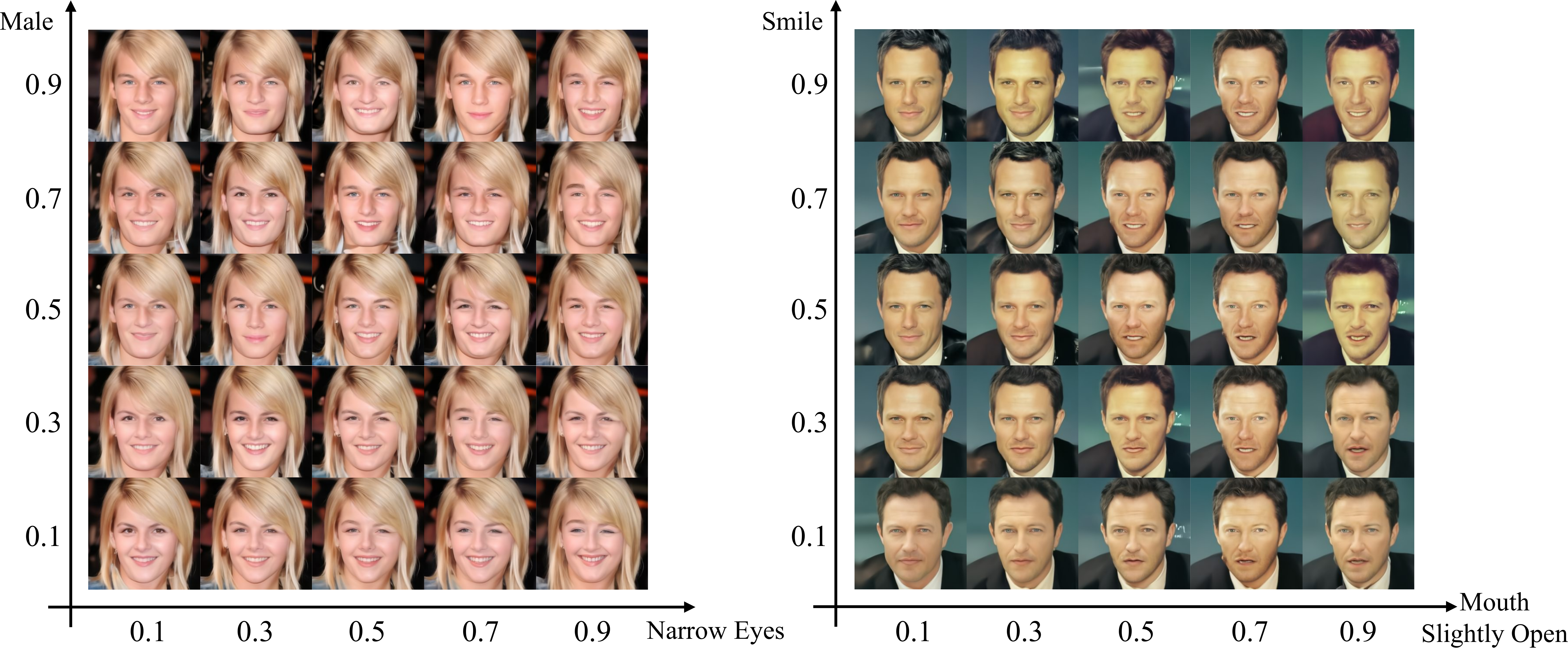}
    \caption{Additional examples demonstrating customizable de-identification. Images are de-identified with the attribute values in the source attribute distribution for `Smile' and `Mouth Slightly Open', and `Male' and `Narrow eyes' fixed from 0.1 to 0.9.}
    \label{sup:smileopen1}
\end{figure*}

\subsection{Visualizing the Identity-Utility Trade-off}
Although the privacy-utility trade-off is quantitatively measured by offering a nested harmonic mean, we additionally visualize this trade-off by presenting a Pareto front curve for a more comprehensive and fair comparison. We select two pillars from Table \ref{tab:comparison}; SID and the harmonic mean of three attributes. We do not include detectability as the third pillar because the difference is trivial across all models (up to $0.2\%$). The Pareto front curve highlights models that achieve an optimal balance between these competing objectives, in the sense that no model on the curve can improve one pillar without degrading the other. Models lying below the curve are strictly dominated, indicating inferior trade-offs compared to those on the Pareto frontier. As displayed in Figure \ref{fig:pareto}, FaceNet-based FLUID and the linear variant of MagFace form a Pareto front in CelebA-HQ, while all FLUID results are included in the Pareto front, highlighting the superior balance between identity removal and utility preservation quantitatively. Meanwhile, the balance of FAMS differs significantly across two datasets due to the lowest gender and age preservation scores at CelebA-HQ. FALCO and DeepPrivacy2 locate far from the Pareto front, which is supported by the lowest SID scores.

\subsection{Customizable de-identification.}
In many real-world applications, complete attribute preservation during de-identification may not be desirable or necessary. For instance, medical researchers studying facial expressions in autism spectrum disorders may need to preserve emotional attributes while removing identity-related features. Conversely, demographic studies might require preserving gender attributes while obscuring other characteristics. This necessitates a flexible approach where practitioners can selectively choose which attributes to modify during the de-identification process.

We modify the behavior of the attribute loss by enabling and partially fixing components of the entire 40 attribute distribution. While our method originally targeted only two attributes from the full distribution of the source image—thereby preserving gender and age—we now assign varying probability values to specific attributes, ranging from 0.1 to 0.9, to explore targeted manipulations along with de-identification. Results of changing \{`Male', `Narrow Eyes'\} and \{`Smile', `Mouth Slightly Open'\} are shown in Figure \ref{sup:smileopen1}. When attributes `Smile' and `Mouth Slightly Open' are assigned low probabilities, the de-identified faces appear neutral. A high probability for `Mouth Slightly Open' and a low one for `Smile' produce an emotionless face with an open mouth. Conversely, assigning a high probability to `Smile' and a low one to `Mouth Slightly Open' generates a closed-mouth smile. High probabilities for both yield a grin. Adjusting the probabilities of `Male' and `Narrow Eyes' also produces interpolated effects. This supports our claim that customizable de-identification can be realized without training the generative model.

\begin{figure}[t]
    \centering
    \includegraphics[width=0.6\columnwidth]{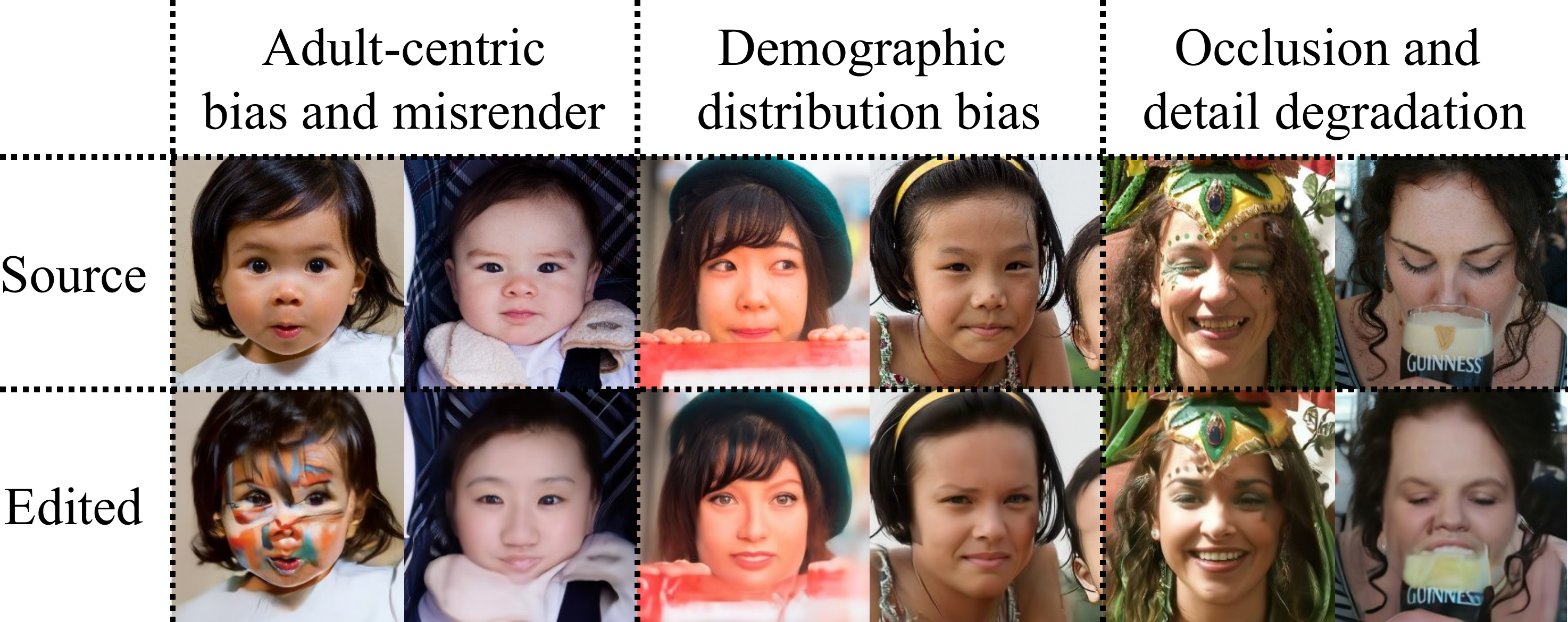}
    \caption{Failure examples and biases from out-of-domain FFHQ edit results.}
    \label{sup:failure}
\end{figure}

\subsection{Failure Examples and Dataset Bias}

Figure \ref{sup:failure} shows representative failure cases in which our method produces noticeable quality degradation and artifacts on challenging inputs. These failures occur primarily for images that lie outside the CelebA‑HQ training distribution of the DM. First, the adult-centric generation tendency renders infant faces with adult‑like features or severe distortions. Second, the demographic distribution bias occasionally shifts facial characteristics and geometry toward stereotypical White ethnicity, effectively `whitening' the minority faces. Finally, faces with heavy occlusions and fine details, such as glasses or jewelry, are frequently simplified or frontalized, losing important structural cues. These limitations reflect both the celebrity‑centric biases of the training data and the inherent smoothing effects of diffusion representations \cite{prabhu2019covering, munoz2023uncovering}. Although our method delivers competitive performance, these failure modes stress the challenges when extending to more diverse, real‑world inputs.

\section{Limitations and Societal Impact}
\label{sec:limitations}
While our method excels in identity suppression, attribute preservation, and visual quality, it has some limitations. Its iterative latent-space optimization hinders real-time use in resource-limited settings. Performance may drop on highly out-of-distribution data, given its design assumptions and reliance on CelebA-HQ-style priors. Also it may not reliably preserve high-frequency details (e.g., text, glasses, occlusions). Finally, white-box attack against angular distance-based FR models such as MagFace with a more sophisticated design will result in higher identity-utility balances.

\section{Conclusion}
We introduced FLUID, a novel training-free framework for face de-identification via identity substitution in the latent \textit{h}-space of pretrained DMs. By leveraging disentangled reagent losses, FLUID enables fine-grained identity removal while preserving facial semantics and visual fidelity. We proposed and empirically supported both linear and geodesic (tangent-based) editing strategies, grounded by theoretical justification. Extensive experiments on CelebA-HQ and FFHQ demonstrate that FLUID achieves competitive de-identification performance with outperforming balance across identity suppression, attribute preservation, and image quality, all without requiring additional training or auxiliary data. This work opens promising directions for controllable, explainable, and privacy-preserving face editing, with practical implications for secure data sharing, ethical AI applications, and real-world de-identification systems.

\bibliographystyle{ACM-Reference-Format}
\bibliography{main}

@String{Computing = "Computing" }

@String{Computer = "{IEEE} Computer" }

@String{Springer = "Springer-Verlag" }

@article{ho2020denoising,
  title={Denoising diffusion probabilistic models},
  author={Ho, Jonathan and Jain, Ajay and Abbeel, Pieter},
  journal={Advances in neural information processing systems},
  volume={33},
  pages={6840--6851},
  year={2020}
}

@inproceedings{ilia2015face,
  title={Face/off: Preventing privacy leakage from photos in social networks},
  author={Ilia, Panagiotis and Polakis, Iasonas and Athanasopoulos, Elias and Maggi, Federico and Ioannidis, Sotiris},
  booktitle={Proceedings of the 22nd ACM SIGSAC Conference on computer and communications security},
  pages={781--792},
  year={2015}
}

@article{babaguchi2009psychological,
  title={Psychological study for designing privacy protected video surveillance system: PriSurv},
  author={Babaguchi, Noboru and Koshimizu, Takashi and Umata, Ichiro and Toriyama, Tomoji},
  journal={Protecting Privacy in Video Surveillance},
  pages={147--164},
  year={2009},
  publisher={Springer}
}

@article{goodfellow2014generative,
  title={Generative adversarial nets},
  author={Goodfellow, Ian J and Pouget-Abadie, Jean and Mirza, Mehdi and Xu, Bing and Warde-Farley, David and Ozair, Sherjil and Courville, Aaron and Bengio, Yoshua},
  journal={Advances in neural information processing systems},
  volume={27},
  year={2014}
}

@inproceedings{karras2019style,
  title={A style-based generator architecture for generative adversarial networks},
  author={Karras, Tero and Laine, Samuli and Aila, Timo},
  booktitle={Proceedings of the IEEE/CVF conference on computer vision and pattern recognition},
  pages={4401--4410},
  year={2019}
}

@inproceedings{karras2020analyzing,
  title={Analyzing and improving the image quality of stylegan},
  author={Karras, Tero and Laine, Samuli and Aittala, Miika and Hellsten, Janne and Lehtinen, Jaakko and Aila, Timo},
  booktitle={Proceedings of the IEEE/CVF conference on computer vision and pattern recognition},
  pages={8110--8119},
  year={2020}
}

@article{kung2025nullface,
  title={NullFace: Training-Free Localized Face Anonymization},
  author={Kung, Han-Wei and Varanka, Tuomas and Sim, Terence and Sebe, Nicu},
  journal={arXiv preprint arXiv:2503.08478},
  year={2025}
}

@inproceedings{sun2024diffam,
  title={Diffam: Diffusion-based adversarial makeup transfer for facial privacy protection},
  author={Sun, Yuhao and Yu, Lingyun and Xie, Hongtao and Li, Jiaming and Zhang, Yongdong},
  booktitle={Proceedings of the IEEE/CVF conference on computer vision and pattern recognition},
  pages={24584--24594},
  year={2024}
}

@article{le2025diffprivate,
  title={DiffPrivate: Facial Privacy Protection with Diffusion Models},
  author={Le, Minh-Ha and Carlsson, Niklas},
  journal={Proceedings on Privacy Enhancing Technologies},
  year={2025}
}

@article{he2024diff,
  title={Diff-privacy: Diffusion-based face privacy protection},
  author={He, Xiao and Zhu, Mingrui and Chen, Dongxin and Wang, Nannan and Gao, Xinbo},
  journal={IEEE Transactions on Circuits and Systems for Video Technology},
  year={2024},
  publisher={IEEE}
}

@inproceedings{liu2024adv,
  title={Adv-diffusion: imperceptible adversarial face identity attack via latent diffusion model},
  author={Liu, Decheng and Wang, Xijun and Peng, Chunlei and Wang, Nannan and Hu, Ruimin and Gao, Xinbo},
  booktitle={Proceedings of the AAAI conference on artificial intelligence},
  volume={38},
  number={4},
  pages={3585--3593},
  year={2024}
}

@inproceedings{lugmayr2022repaint,
  title={Repaint: Inpainting using denoising diffusion probabilistic models},
  author={Lugmayr, Andreas and Danelljan, Martin and Romero, Andres and Yu, Fisher and Timofte, Radu and Van Gool, Luc},
  booktitle={Proceedings of the IEEE/CVF conference on computer vision and pattern recognition},
  pages={11461--11471},
  year={2022}
}

@inproceedings{shaheryar2024iddiffuse,
  title={IDDiffuse: Dual-Conditional Diffusion Model for Enhanced Facial Image Anonymization},
  author={Shaheryar, Muhammad and Lee, Jong Taek and Jung, Soon Ki},
  booktitle={Proceedings of the Asian Conference on Computer Vision},
  pages={4017--4033},
  year={2024}
}

@article{wen2022identitydp,
  title={Identitydp: Differential private identification protection for face images},
  author={Wen, Yunqian and Liu, Bo and Ding, Ming and Xie, Rong and Song, Li},
  journal={Neurocomputing},
  volume={501},
  pages={197--211},
  year={2022},
  publisher={Elsevier}
}

@inproceedings{kung2025face,
  title={Face anonymization made simple},
  author={Kung, Han-Wei and Varanka, Tuomas and Saha, Sanjay and Sim, Terence and Sebe, Nicu},
  booktitle={2025 IEEE/CVF Winter Conference on Applications of Computer Vision (WACV)},
  pages={1040--1050},
  year={2025},
  organization={IEEE}
}

@article{dhariwal2021diffusion,
  title={Diffusion models beat gans on image synthesis},
  author={Dhariwal, Prafulla and Nichol, Alexander},
  journal={Advances in neural information processing systems},
  volume={34},
  pages={8780--8794},
  year={2021}
}

@inproceedings{stypulkowski2024diffused,
  title={Diffused heads: Diffusion models beat gans on talking-face generation},
  author={Stypu{\l}kowski, Micha{\l} and Vougioukas, Konstantinos and He, Sen and Zi{\k{e}}ba, Maciej and Petridis, Stavros and Pantic, Maja},
  booktitle={Proceedings of the IEEE/CVF Winter Conference on Applications of Computer Vision},
  pages={5091--5100},
  year={2024}
}

@inproceedings{kim2022diffusionclip,
  title={Diffusionclip: Text-guided diffusion models for robust image manipulation},
  author={Kim, Gwanghyun and Kwon, Taesung and Ye, Jong Chul},
  booktitle={Proceedings of the IEEE/CVF conference on computer vision and pattern recognition},
  pages={2426--2435},
  year={2022}
}

@article{song2020denoising,
  title={Denoising diffusion implicit models},
  author={Song, Jiaming and Meng, Chenlin and Ermon, Stefano},
  journal={arXiv preprint arXiv:2010.02502},
  year={2020}
}

@inproceedings{avrahami2022blended,
  title={Blended diffusion for text-driven editing of natural images},
  author={Avrahami, Omri and Lischinski, Dani and Fried, Ohad},
  booktitle={Proceedings of the IEEE/CVF conference on computer vision and pattern recognition},
  pages={18208--18218},
  year={2022}
}

@inproceedings{hu2022protecting,
  title={Protecting facial privacy: Generating adversarial identity masks via style-robust makeup transfer},
  author={Hu, Shengshan and Liu, Xiaogeng and Zhang, Yechao and Li, Minghui and Zhang, Leo Yu and Jin, Hai and Wu, Libing},
  booktitle={Proceedings of the IEEE/CVF conference on computer vision and pattern recognition},
  pages={15014--15023},
  year={2022}
}

@inproceedings{shamshad2023clip2protect,
  title={Clip2protect: Protecting facial privacy using text-guided makeup via adversarial latent search},
  author={Shamshad, Fahad and Naseer, Muzammal and Nandakumar, Karthik},
  booktitle={Proceedings of the IEEE/CVF Conference on Computer Vision and Pattern Recognition},
  pages={20595--20605},
  year={2023}
}

@inproceedings{shan2020fawkes,
  title={Fawkes: Protecting privacy against unauthorized deep learning models},
  author={Shan, Shawn and Wenger, Emily and Zhang, Jiayun and Li, Huiying and Zheng, Haitao and Zhao, Ben Y},
  booktitle={29th USENIX security symposium (USENIX Security 20)},
  pages={1589--1604},
  year={2020}
}

@article{yang2024g,
  title={G 2 face: High-fidelity reversible face anonymization via generative and geometric priors},
  author={Yang, Haoxin and Xu, Xuemiao and Xu, Cheng and Zhang, Huaidong and Qin, Jing and Wang, Yi and Heng, Pheng-Ann and He, Shengfeng},
  journal={IEEE Transactions on Information Forensics and Security},
  year={2024},
  publisher={IEEE}
}

@inproceedings{barattin2023attribute,
  title={Attribute-preserving face dataset anonymization via latent code optimization},
  author={Barattin, Simone and Tzelepis, Christos and Patras, Ioannis and Sebe, Nicu},
  booktitle={Proceedings of the IEEE/CVF conference on computer vision and pattern recognition},
  pages={8001--8010},
  year={2023}
}

@article{kim2025visual,
  title={Visual context-aware attribute-preserving face de-identification},
  author={Kim, Hyeonwoo and Shim, Jonghwa and Park, Sungwoo and Hwang, Eenjun},
  journal={Neurocomputing},
  volume={638},
  pages={130205},
  year={2025},
  publisher={Elsevier}
}

@inproceedings{maximov2020ciagan,
  title={Ciagan: Conditional identity anonymization generative adversarial networks},
  author={Maximov, Maxim and Elezi, Ismail and Leal-Taix{\'e}, Laura},
  booktitle={Proceedings of the IEEE/CVF conference on computer vision and pattern recognition},
  pages={5447--5456},
  year={2020}
}

@inproceedings{hukkelaas2019deepprivacy,
  title={Deepprivacy: A generative adversarial network for face anonymization},
  author={Hukkel{\aa}s, H{\aa}kon and Mester, Rudolf and Lindseth, Frank},
  booktitle={International symposium on visual computing},
  pages={565--578},
  year={2019},
  organization={Springer}
}

@inproceedings{hukkelaas2023deepprivacy2,
  title={Deepprivacy2: Towards realistic full-body anonymization},
  author={Hukkel{\aa}s, H{\aa}kon and Lindseth, Frank},
  booktitle={Proceedings of the IEEE/CVF winter conference on applications of computer vision},
  pages={1329--1338},
  year={2023}
}

@article{li2023riddle,
  title={Riddle: Reversible and diversified de-identification with latent encryptor},
  author={Li, Dongze and Wang, Wei and Zhao, Kang and Dong, Jing and Tan, Tieniu},
  journal={arXiv preprint arXiv:2303.05171},
  year={2023}
}

@inproceedings{cai2024disguise,
  title={Disguise without disruption: Utility-preserving face de-identification},
  author={Cai, Zikui and Gao, Zhongpai and Planche, Benjamin and Zheng, Meng and Chen, Terrence and Asif, M Salman and Wu, Ziyan},
  booktitle={Proceedings of the AAAI Conference on Artificial Intelligence},
  volume={38},
  number={2},
  pages={918--926},
  year={2024}
}

@article{kwon2022diffusion,
  title={Diffusion models already have a semantic latent space},
  author={Kwon, Mingi and Jeong, Jaeseok and Uh, Youngjung},
  journal={arXiv preprint arXiv:2210.10960},
  year={2022}
}

@article{zhu2023boundary,
  title={Boundary guided learning-free semantic control with diffusion models},
  author={Zhu, Ye and Wu, Yu and Deng, Zhiwei and Russakovsky, Olga and Yan, Yan},
  journal={Advances in Neural Information Processing Systems},
  volume={36},
  pages={78319--78346},
  year={2023}
}

@inproceedings{jeong2024training,
  title={Training-free content injection using h-space in diffusion models},
  author={Jeong, Jaeseok and Kwon, Mingi and Uh, Youngjung},
  booktitle={Proceedings of the IEEE/CVF Winter Conference on Applications of Computer Vision},
  pages={5151--5161},
  year={2024}
}

@article{harkonen2020ganspace,
  title={Ganspace: Discovering interpretable gan controls},
  author={H{\"a}rk{\"o}nen, Erik and Hertzmann, Aaron and Lehtinen, Jaakko and Paris, Sylvain},
  journal={Advances in neural information processing systems},
  volume={33},
  pages={9841--9850},
  year={2020}
}

@article{shen2020interfacegan,
  title={Interfacegan: Interpreting the disentangled face representation learned by gans},
  author={Shen, Yujun and Yang, Ceyuan and Tang, Xiaoou and Zhou, Bolei},
  journal={IEEE transactions on pattern analysis and machine intelligence},
  volume={44},
  number={4},
  pages={2004--2018},
  year={2020},
  publisher={IEEE}
}

@article{abdal2021styleflow,
  title={Styleflow: Attribute-conditioned exploration of stylegan-generated images using conditional continuous normalizing flows},
  author={Abdal, Rameen and Zhu, Peihao and Mitra, Niloy J and Wonka, Peter},
  journal={ACM Transactions on Graphics (ToG)},
  volume={40},
  number={3},
  pages={1--21},
  year={2021},
  publisher={ACM New York, NY}
}

@inproceedings{wu2021stylespace,
  title={Stylespace analysis: Disentangled controls for stylegan image generation},
  author={Wu, Zongze and Lischinski, Dani and Shechtman, Eli},
  booktitle={Proceedings of the IEEE/CVF conference on computer vision and pattern recognition},
  pages={12863--12872},
  year={2021}
}

@inproceedings{haas2024discovering,
  title={Discovering interpretable directions in the semantic latent space of diffusion models},
  author={Haas, Ren{\'e} and Huberman-Spiegelglas, Inbar and Mulayoff, Rotem and Gra{\ss}hof, Stella and Brandt, Sami S and Michaeli, Tomer},
  booktitle={2024 IEEE 18th International Conference on Automatic Face and Gesture Recognition (FG)},
  pages={1--9},
  year={2024},
  organization={IEEE}
}

@inproceedings{kim2023dcface,
  title={Dcface: Synthetic face generation with dual condition diffusion model},
  author={Kim, Minchul and Liu, Feng and Jain, Anil and Liu, Xiaoming},
  booktitle={Proceedings of the ieee/cvf conference on computer vision and pattern recognition},
  pages={12715--12725},
  year={2023}
}

@inproceedings{zhao2023diffswap,
  title={Diffswap: High-fidelity and controllable face swapping via 3d-aware masked diffusion},
  author={Zhao, Wenliang and Rao, Yongming and Shi, Weikang and Liu, Zuyan and Zhou, Jie and Lu, Jiwen},
  booktitle={Proceedings of the IEEE/CVF Conference on Computer Vision and Pattern Recognition},
  pages={8568--8577},
  year={2023}
}

@article{westerlund2019emergence,
  title={The emergence of deepfake technology: A review},
  author={Westerlund, Mika},
  journal={Technology innovation management review},
  volume={9},
  number={11},
  year={2019}
}

@inproceedings{ronneberger2015u,
  title={U-net: Convolutional networks for biomedical image segmentation},
  author={Ronneberger, Olaf and Fischer, Philipp and Brox, Thomas},
  booktitle={Medical image computing and computer-assisted intervention--MICCAI 2015: 18th international conference, Munich, Germany, October 5-9, 2015, proceedings, part III 18},
  pages={234--241},
  year={2015},
  organization={Springer}
}

@inproceedings{schroff2015facenet,
  title={Facenet: A unified embedding for face recognition and clustering},
  author={Schroff, Florian and Kalenichenko, Dmitry and Philbin, James},
  booktitle={Proceedings of the IEEE conference on computer vision and pattern recognition},
  pages={815--823},
  year={2015}
}

@inproceedings{deng2019arcface,
  title={Arcface: Additive angular margin loss for deep face recognition},
  author={Deng, Jiankang and Guo, Jia and Xue, Niannan and Zafeiriou, Stefanos},
  booktitle={Proceedings of the IEEE/CVF conference on computer vision and pattern recognition},
  pages={4690--4699},
  year={2019}
}

@inproceedings{serengil2021hyperextended,
  title={Hyperextended lightface: A facial attribute analysis framework},
  author={Serengil, Sefik Ilkin and Ozpinar, Alper},
  booktitle={2021 International Conference on Engineering and Emerging Technologies (ICEET)},
  pages={1--4},
  year={2021},
  organization={IEEE}
}

@inproceedings{liu2017sphereface,
  title={Sphereface: Deep hypersphere embedding for face recognition},
  author={Liu, Weiyang and Wen, Yandong and Yu, Zhiding and Li, Ming and Raj, Bhiksha and Song, Le},
  booktitle={Proceedings of the IEEE conference on computer vision and pattern recognition},
  pages={212--220},
  year={2017}
}

@inproceedings{chen2018mobilefacenets,
  title={Mobilefacenets: Efficient cnns for accurate real-time face verification on mobile devices},
  author={Chen, Sheng and Liu, Yang and Gao, Xiang and Han, Zhen},
  booktitle={Chinese conference on biometric recognition},
  pages={428--438},
  year={2018},
  organization={Springer}
}

@article{qin2023swinface,
  title={SwinFace: A multi-task transformer for face recognition, expression recognition, age estimation and attribute estimation},
  author={Qin, Lixiong and Wang, Mei and Deng, Chao and Wang, Ke and Chen, Xi and Hu, Jiani and Deng, Weihong},
  journal={IEEE Transactions on Circuits and Systems for Video Technology},
  volume={34},
  number={4},
  pages={2223--2234},
  year={2023},
  publisher={IEEE}
}

@article{narayan2024facexformer,
  title={Facexformer: A unified transformer for facial analysis},
  author={Narayan, Kartik and VS, Vibashan and Chellappa, Rama and Patel, Vishal M},
  journal={arXiv preprint arXiv:2403.12960},
  year={2024}
}

@article{karras2017progressive,
  title={Progressive Growing of GANs for Improved Quality, Stability, and Variation},
  author={Karras, Tero},
  journal={arXiv preprint arXiv:1710.10196},
  year={2017}
}

@inproceedings{liu2015deep,
  title={Deep learning face attributes in the wild},
  author={Liu, Ziwei and Luo, Ping and Wang, Xiaogang and Tang, Xiaoou},
  booktitle={Proceedings of the IEEE international conference on computer vision},
  pages={3730--3738},
  year={2015}
}

@inproceedings{terhorst2020ser,
  title={SER-FIQ: Unsupervised estimation of face image quality based on stochastic embedding robustness},
  author={Terhorst, Philipp and Kolf, Jan Niklas and Damer, Naser and Kirchbuchner, Florian and Kuijper, Arjan},
  booktitle={Proceedings of the IEEE/CVF conference on computer vision and pattern recognition},
  pages={5651--5660},
  year={2020}
}

@article{kingma2014adam,
  title={Adam: A method for stochastic optimization},
  author={Kingma, Diederik P and Ba, Jimmy},
  journal={arXiv preprint arXiv:1412.6980},
  year={2014}
}

@article{hahm2024isometric,
  title={Isometric representation learning for disentangled latent space of diffusion models},
  author={Hahm, Jaehoon and Lee, Junho and Kim, Sunghyun and Lee, Joonseok},
  journal={arXiv preprint arXiv:2407.11451},
  year={2024}
}

@misc{gdpr,
  title = {Regulation (EU) 2016/679 (General Data Protection Regulation)},
  author = {{European Union}},
  year = {2016},
  note = {Official Journal of the European Union, L119, 1–88},
  url = {https://eur-lex.europa.eu/eli/reg/2016/679/oj/}
}

@misc{dpdpa,
  title = {Digital Personal Data Protection Act, 2023},
  author = {{Government of India}},
  year = {2023},
  note = {Ministry of Law and Justice},
  url = {https://www.meity.gov.in/static/uploads/2024/06/2bf1f0e9f04e6fb4f8fef35e82c42aa5.pdf}
}

@inproceedings{meng2021magface,
  title={Magface: A universal representation for face recognition and quality assessment},
  author={Meng, Qiang and Zhao, Shichao and Huang, Zhida and Zhou, Feng},
  booktitle={Proceedings of the IEEE/CVF conference on computer vision and pattern recognition},
  pages={14225--14234},
  year={2021}
}

@inproceedings{shaheryar2025black,
  title={Black Hole-Driven Identity Absorbing in Diffusion Models},
  author={Shaheryar, Muhammad and Lee, Jong Taek and Jung, Soon Ki},
  booktitle={Proceedings of the Computer Vision and Pattern Recognition Conference},
  pages={28544--28554},
  year={2025}
}

@article{chen2016training,
  title={Training deep nets with sublinear memory cost},
  author={Chen, Tianqi and Xu, Bing and Zhang, Chiyuan and Guestrin, Carlos},
  journal={arXiv preprint arXiv:1604.06174},
  year={2016}
}

@inproceedings{choi2022perception,
  title={Perception prioritized training of diffusion models},
  author={Choi, Jooyoung and Lee, Jungbeom and Shin, Chaehun and Kim, Sungwon and Kim, Hyunwoo and Yoon, Sungroh},
  booktitle={Proceedings of the IEEE/CVF conference on computer vision and pattern recognition},
  pages={11472--11481},
  year={2022}
}

@article{prabhu2019covering,
  title={Covering up bias in CelebA-like datasets with Markov blankets: A post-hoc cure for attribute prior avoidance},
  author={Prabhu, Vinay Uday and Yap, Dian Ang and Wang, Alexander and Whaley, John},
  journal={arXiv preprint arXiv:1907.12917},
  year={2019}
}

@article{munoz2023uncovering,
  title={Uncovering bias in face generation models},
  author={Mu{\~n}oz, Cristian and Zannone, Sara and Mohammed, Umar and Koshiyama, Adriano},
  journal={arXiv preprint arXiv:2302.11562},
  year={2023}
}

@article{king2009dlib,
  title={Dlib-ml: A machine learning toolkit},
  author={King, Davis E},
  journal={The Journal of Machine Learning Research},
  volume={10},
  pages={1755--1758},
  year={2009},
  publisher={JMLR. org}
}

@article{zhang2016joint,
  title={Joint face detection and alignment using multitask cascaded convolutional networks},
  author={Zhang, Kaipeng and Zhang, Zhanpeng and Li, Zhifeng and Qiao, Yu},
  journal={IEEE signal processing letters},
  volume={23},
  number={10},
  pages={1499--1503},
  year={2016},
  publisher={IEEE}
}

@article{heusel2017gans,
  title={Gans trained by a two time-scale update rule converge to a local nash equilibrium},
  author={Heusel, Martin and Ramsauer, Hubert and Unterthiner, Thomas and Nessler, Bernhard and Hochreiter, Sepp},
  journal={Advances in neural information processing systems},
  volume={30},
  year={2017}
}

@inproceedings{szegedy2015going,
  title={Going deeper with convolutions},
  author={Szegedy, Christian and Liu, Wei and Jia, Yangqing and Sermanet, Pierre and Reed, Scott and Anguelov, Dragomir and Erhan, Dumitru and Vanhoucke, Vincent and Rabinovich, Andrew},
  booktitle={Proceedings of the IEEE conference on computer vision and pattern recognition},
  pages={1--9},
  year={2015}
}

@inproceedings{deng2009imagenet,
  title={Imagenet: A large-scale hierarchical image database},
  author={Deng, Jia and Dong, Wei and Socher, Richard and Li, Li-Jia and Li, Kai and Fei-Fei, Li},
  booktitle={2009 IEEE conference on computer vision and pattern recognition},
  pages={248--255},
  year={2009},
  organization={Ieee}
}

@inproceedings{pan2023effective,
  title={Effective real image editing with accelerated iterative diffusion inversion},
  author={Pan, Zhihong and Gherardi, Riccardo and Xie, Xiufeng and Huang, Stephen},
  booktitle={Proceedings of the IEEE/CVF International Conference on Computer Vision},
  pages={15912--15921},
  year={2023}
}

@inproceedings{radiya-dixit2022data,
  title={DATA POISONING WON’T SAVE YOU FROM FACIAL RECOGNITION},
  author={Radiya-Dixit, Evani and Hong, Sanghyun and Carlini, Nicholas and Tram{\`e}r, Florian},
  booktitle={International Conference on Learning Representations (ICLR)},
  year={2022}
}

@inproceedings{oh2016faceless,
  title={Faceless person recognition: Privacy implications in social media},
  author={Oh, Seong Joon and Benenson, Rodrigo and Fritz, Mario and Schiele, Bernt},
  booktitle={European Conference on Computer Vision},
  pages={19--35},
  year={2016},
  organization={Springer}
}

@article{nie2022diffusion,
  title={Diffusion models for adversarial purification},
  author={Nie, Weili and Guo, Brandon and Huang, Yujia and Xiao, Chaowei and Vahdat, Arash and Anandkumar, Anima},
  journal={arXiv preprint arXiv:2205.07460},
  year={2022}
}

\end{document}